\documentclass{dexgem-report}
\usepackage{amsmath,amssymb}
\usepackage{pifont}
\usepackage{float}
\usepackage{listings}
\usepackage{tikz}
\usetikzlibrary{arrows.meta,positioning,calc}

\newcommand{\xmark}{\ding{55}}
\definecolor{codeback}{HTML}{F5F7FA}
\title{DexCanvas: Bridging Human Demonstrations and Robot Learning for Dexterous Manipulation}
\reportdate{October 2025}

\author{Xinyue Xu\affil{2,3}\textsuperscript{*} \and Jieqiang Sun\affil{1}\textsuperscript{*} \and Jing (Daisy) Dai\affil{2,4}\textsuperscript{*} \and Siyuan Chen\affil{1} \and Lanjie Ma\affil{2,5} \and Ke Sun\affil{2} \and Bin Zhao\affil{1,6} \and Jianbo Yuan\affil{2,4} \and Sheng Yi\affil{1,2} \and Haohua Zhu\affil{1,2} \and Irvine Lu\affil{1}\textsuperscript{\ensuremath{\dagger}}}
\affiliation{1}{DexGEM Lab}
\affiliation{2}{DexRobot Co.\ Ltd.}
\affiliation{3}{University of Michigan}
\affiliation{4}{Shanghai Jiao Tong University}
\affiliation{5}{Chongqing University}
\affiliation{6}{East China University of Science and Technology}
\authornote{\textsuperscript{*}Equal contribution.\quad \textsuperscript{\ensuremath{\dagger}}Corresponding author: Irvine Lu (\href{mailto:irvine@dex-gem.ai}{irvine@dex-gem.ai}).}
\datalink{https://huggingface.co/datasets/Dex-GEM/DexCanvas}

\begin{document}
\maketitle

\begin{abstract}
A dexterous hand succeeds or fails through contact, yet every scalable record of human manipulation (video, vision-based hand tracking, motion capture) observes how the hand moves and misses how it pushes. We present DexCanvas, a hybrid real--synthetic dataset of 7{,}000 hours of dexterous hand--object interaction with per-frame contact points and contact forces, grown from 70 hours of optical motion capture that spans 21 manipulation types, from power and precision grasps to in-hand manipulation. The forces come from a real-to-sim pipeline: one reinforcement learning (RL) policy, conditioned on object geometry and manipulation type, drives an actuated MANO hand in physics simulation to reproduce each captured object trajectory, and the simulator measures the contacts that produce the reproduced motion. The policy reproduces 87.5\% of the demonstrations and still succeeds on 80.9\% of rollouts from perturbed initial object poses; these perturbed rollouts expand the 70 captured hours a hundredfold. On six YCB objects absent from the capture set it succeeds 72.5\% of the time. The contact channels make better policies: a Diffusion Policy that observes DexCanvas contact features succeeds 82.2\% of the time, against 67.5\% without them, over 18{,}522 trials. Contact-driven retargeting carries the same demonstrations to two dexterous robot hands, where the RL stage again produces stable rollouts. A first release of 111{,}857 force-annotated trajectories is publicly available on Hugging Face.
\end{abstract}

\section{Introduction}

Anthropomorphic hands are the most general interface a robot can have with the physical world, and they are a prerequisite for human-level autonomy in unstructured environments~\citep{yu2022dexterous,ozawa2017grasp}. Robot learning now offers the machinery to train such hands: reinforcement learning for contact-rich control~\citep{objdex2024,bidexhands2024}, diffusion models for multimodal action distributions~\citep{dexdiffuser2024,unidexfpm2024}, and foundation models with billions of parameters~\citep{dexvla2025,openvla2024,rt2-2024}. Yet the flagship manipulation systems built on this machinery still grasp with parallel-jaw grippers, and generalizable control of multi-fingered hands remains confined to simulation or narrow real-world settings. The missing ingredient is data: dexterous control needs demonstrations that record both the diverse strategies humans use and the contact forces that make those strategies work.

Every available data source supplies one of these and lacks the other. Teleoperation~\citep{arunachalam2022dexterous} records robot actions on the target hardware, but it is slow and expensive, and controlling a high-DoF hand imposes a cognitive load that produces unnatural finger motion. Synthetic generation scales to millions of grasps through optimization~\citep{wang2022dexgraspnet,zhang2024dexgraspnet} and to billions through generative models~\citep{dex1b2025}, but its grasps are often biomechanically implausible and are tied to one robot morphology. Human demonstrations are natural and morphology-agnostic. Internet-scale and egocentric video~\citep{Hoque2025EgoDex,damen2018epic} offers breadth without systematic coverage; vision-based capture~\citep{Chao2021DexYCB,hampali2020honnotate,dexmv2022} suffers from self-occlusion and noisy 3D pose; motion capture~\citep{Fan2023ARCTIC,Taheri2020GRAB,wang2024dexcap} delivers precise kinematics. All of them record geometry and none records force.

DexCanvas starts from one observation: forces need not be measured if they can be reconstructed. When a physically simulated hand reproduces the object motion of a human demonstration, the contact forces it exerts are, by construction, forces that produce that motion under physics. Reconstruction thus turns a kinematic recording into a dynamic one, and once a policy can reproduce a demonstration, it can also reproduce it from many perturbed starting conditions, each yielding a new physically valid trajectory.

\begin{figure}[t]
    \centering
    \includegraphics[width=\linewidth]{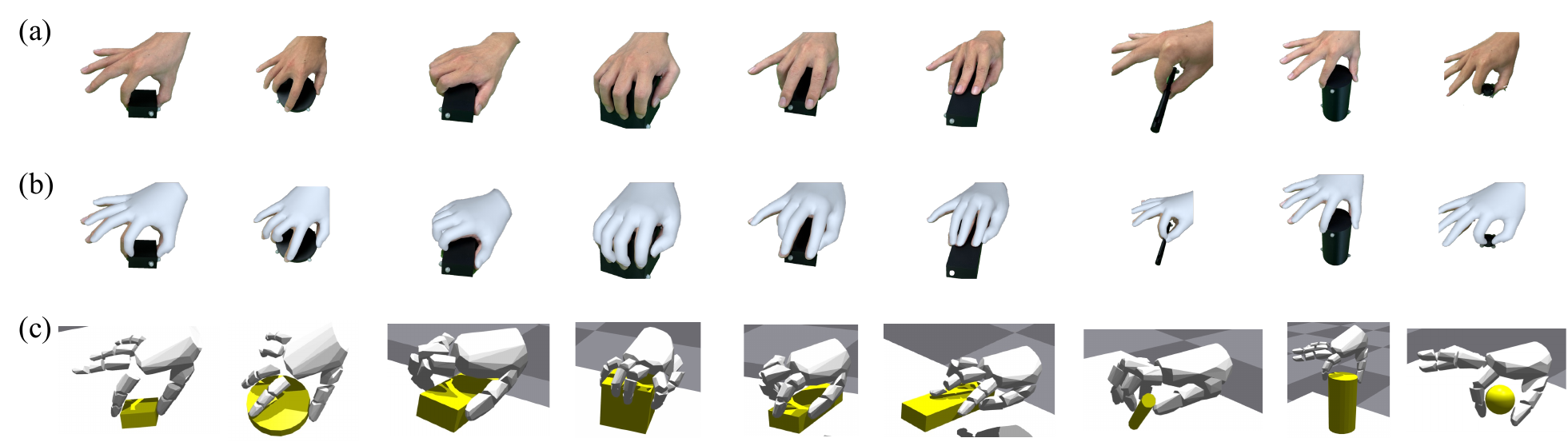}
    \caption{From human demonstration to force-annotated trajectory. (a) Human manipulation captured with optical motion capture across diverse manipulation types. (b) MANO hand fitted to the motion-capture markers. (c) An actuated MANO hand reproduces the demonstration in physics simulation; the simulator records the contact forces that generate the observed object motion.}
    \label{fig:teaser_figure}
\end{figure}

We build DexCanvas\footnote{The name casts the dataset as a foundational ``canvas'' for dexterous manipulation, with human demonstrations as the raw material from which robot skills are drawn.} on this observation (Figure~\ref{fig:teaser_figure}). We record 70 hours of human manipulation (12{,}000 sequences) with a 22-camera optical motion-capture system and two synchronized RGB-D cameras, organized around 21 manipulation types that combine grasp types from established taxonomies~\citep{feix2016grasp,chen2025dexonomy} with in-hand manipulation, from power grasps through precision pinches to in-hand reorientation. A single RL policy, shared across all object--manipulation pairs, reproduces these demonstrations with an actuated MANO hand~\citep{romero2022embodied} in physics simulation and reads out per-frame contact points and contact forces. Rolling the same policy out from perturbed initial object poses expands the 70 captured hours into 7{,}000 hours of physically validated, force-annotated trajectories. Every trajectory provides MANO shape and pose, wrist and object 6-DoF poses, and per-frame contacts, and links to the camera streams of the capture that seeded it.

The resulting data is accurate, generalizable, and useful. The unified policy reproduces 87.5\% of the demonstrations and succeeds on 80.9\% of perturbed rollouts, and it reaches 72.5\% on six YCB objects that were never captured. A Diffusion Policy that observes DexCanvas contact features raises success from 67.5\% to 82.2\% over 18{,}522 trials. Contact-driven retargeting transfers the trajectories to two dexterous robot hands, DexHand021 and DexHand021Pro, where the same RL stage restores stable execution.

Our contributions are:
\begin{enumerate}
  \item \textbf{DexCanvas}, a hybrid real--synthetic dataset of 7{,}000 hours of human dexterous manipulation over 21 manipulation types that combines motion-capture kinematics, multi-view camera streams, and per-frame contact forces.
  \item \textbf{A unified real-to-sim reconstruction method} in which one residual RL policy reproduces demonstrations across all object--manipulation pairs, extracts their contact forces, synthesizes perturbed variants, and generalizes to unseen objects.
  \item \textbf{Evidence that the data serves robot learning}: contact-aware imitation learning and cross-embodiment retargeting to two dexterous robot hands. The dataset, with hand and object assets and a replay tool, is released on Hugging Face.
\end{enumerate}

\section{Related Work}

\subsection{Human hand--object interaction datasets}

Human hand datasets trace a progression of sensing technology (Table~\ref{tab:dataset_comparison}). Early vision-based work targeted hand pose estimation~\citep{zimmermann2019freihand} and simple grasping~\citep{Chao2021DexYCB}; HO3D~\citep{hampali2020honnotate} and DexYCB achieved multi-view capture at the cost of laborious manual annotation. Motion capture brought kinematic precision: GRAB~\citep{Taheri2020GRAB} records whole-body grasping and ARCTIC~\citep{Fan2023ARCTIC} bimanual manipulation with synchronized RGB-D. Egocentric collections pushed scale: EgoDex~\citep{Hoque2025EgoDex} records 829 hours of tabletop manipulation with native 3D hand tracking on Apple Vision Pro, and OpenEgo~\citep{jawaid2025openego} unifies six datasets with language-aligned action primitives, each exceeding 90 million frames. None of these datasets records force. Proximity-based labels in motion-capture datasets, thermal imaging~\citep{brahmbhatt2019contactdb,brahmbhatt2020contactpose}, and 4D scanning~\citep{zheng2023hi4d} add binary contact without force magnitude, and synthetic hand datasets~\citep{hasson2019learning,li2023renderih,zhang2024hoidiffusion} provide exact annotations for interactions no human performed. DexCanvas pairs real human motion with continuous, physically consistent contact forces at a scale of 7{,}000 hours.

\begin{table}[t]
\centering
\caption{Comparison with representative human hand--object interaction datasets. DexCanvas is the only dataset that pairs real human manipulation with continuous contact forces.}
\label{tab:dataset_comparison}
\footnotesize
\setlength{\tabcolsep}{3pt}
\begin{tabular*}{\linewidth}{@{\extracolsep{\fill}}lcccccc@{}}
\toprule
\textbf{Dataset} & \textbf{Contact} & \textbf{Scope} & \textbf{Physical} & \textbf{Frames} & \textbf{Modality} & \textbf{Labels} \\
\midrule
\multicolumn{7}{@{}l}{\textit{Vision-focused}} \\
FreiHAND~\citep{zimmermann2019freihand} & \xmark & Poses & \xmark & 130K & RGB & Model fit \\
HO3D~\citep{hampali2020honnotate} & \xmark & Poses & \xmark & 78K & RGB-D & Manual \\
DexYCB~\citep{Chao2021DexYCB} & \xmark & Grasp & \xmark & 90K & RGB-D & Manual \\
HOI4D~\citep{hampali2022hoi4d} & \xmark & General & \xmark & 2.4M & RGB-D (ego) & Manual+auto \\
\midrule
\multicolumn{7}{@{}l}{\textit{Motion-focused}} \\
GRAB~\citep{Taheri2020GRAB} & Binary & Grasp & \xmark & 1.6M & Mocap & Markers \\
ARCTIC~\citep{Fan2023ARCTIC} & Binary & Bimanual & \xmark & 2.1M & RGB-D+mocap & Markers \\
Hi4D~\citep{zheng2023hi4d} & Binary & Human$^{a}$ & \xmark & 11K & 4D scans & Multi-view \\
OpenEgo~\citep{jawaid2025openego} & \xmark & General & \xmark & 119.6M & Multimodal & Fusion$^{b}$ \\
EgoDex~\citep{Hoque2025EgoDex} & \xmark & General & \xmark & 90M & RGB+3D & Vision Pro \\
\midrule
\multicolumn{7}{@{}l}{\textit{Synthetic and hybrid}} \\
ObMan~\citep{hasson2019learning} & \xmark & Grasp & \xmark & 150K & Synthetic & Synthesis \\
RenderIH~\citep{li2023renderih} & \xmark & Two-hand & \xmark & 1M & Synthetic & Synthesis \\
HOIDiffusion~\citep{zhang2024hoidiffusion} & \xmark & Grasp & \xmark & N/A$^{c}$ & Hybrid & Diffusion \\
\midrule
\multicolumn{7}{@{}l}{\textit{Contact-aware}} \\
ContactDB~\citep{brahmbhatt2019contactdb} & Binary & Grasp & \xmark & 375K & Thermal & Thermal \\
ContactPose~\citep{brahmbhatt2020contactpose} & Binary & Grasp & \xmark & 2.9M & RGB-D+thermal & Thermal \\
\midrule
\textbf{DexCanvas (ours)} & \textbf{Continuous} & \textbf{General} & \checkmark & \textbf{3.0B} & RGB-D+mocap & Simulation \\
\bottomrule
\end{tabular*}

\vspace{4pt}
{\scriptsize\raggedright \textit{Contact}: force or contact annotation. \textit{Physical}: trajectories validated by physics simulation. \textit{Labels}: annotation source. $^{a}$Human--human interaction. $^{b}$Unified from six datasets. $^{c}$A generation method, not a fixed dataset. \textit{Scope}: Poses = static hand poses; Grasp = grasping; General = diverse manipulation; Bimanual and Two-hand = two hands manipulating or interacting.\par}
\end{table}

\subsection{Synthetic robot manipulation datasets}

Robot-specific synthetic data offers unlimited scale and exact ground truth. DexGraspNet~\citep{wang2022dexgraspnet} optimizes differentiable force closure to produce 1.32 million stable grasps; DexGraspNet 2.0~\citep{zhang2024dexgraspnet} trains a diffusion model on optimized seeds to reach 427 million grasps in cluttered scenes; Dex1B~\citep{dex1b2025} generates one billion demonstrations across several robot hands with a conditional VAE. Further collections target articulated objects~\citep{bao2023dexart}, bimanual coordination~\citep{bidexhands2024}, task-oriented grasping~\citep{zhang2024dextog}, and household task suites~\citep{nasiriany2024robocasa}, building on earlier physics-validated grasp sets~\citep{aktacs2022deep}. Each of these datasets is generated for a specific robot morphology and inherits the grasp distribution of its generator. DexCanvas seeds synthesis with real human demonstrations, so its trajectories carry human manipulation strategies and remain independent of any robot hand until they are retargeted.

\subsection{Contact and force estimation}

Recovering manipulation forces from observation is under-constrained. \citet{pham2015towards} pose it as inverse dynamics: given observed object motion, solve for the hand forces through second-order cone optimization. PressureVision~\citep{grady2022pressurevision} regresses pressure from the appearance changes of soft tissue. Contact-aware reconstruction~\citep{corona2020ganhand,hasson2019learning,zhou2022toch} enforces plausible, temporally consistent hand--object geometry, but geometrically identical contacts can carry very different forces. Instrumented objects and force sensors alter the mechanics they measure, and thermal imaging~\citep{brahmbhatt2019contactdb} records only where contact occurred. DexCanvas resolves the ambiguity with physics: a simulated hand that reproduces the observed motion produces one set of forces that is dynamically consistent with it, for every hand link and every frame.

\subsection{Manipulation taxonomies and coverage}

Grasp taxonomies organize human manipulation into 16 Cutkosky types and the 33 types of the GRASP taxonomy~\citep{feix2016grasp}, yet most datasets cover a handful of them, concentrating on power grasps and simple pinches because pick-and-place dominates robot benchmarks. Dexonomy~\citep{chen2025dexonomy} synthesizes 31 of the 33 GRASP types for robot hands, showing that systematic coverage is achievable. DexCanvas brings the same systematic coverage to real human demonstrations and extends it beyond static grasps to in-hand rolling, sliding, finger transfer, and rotation.

\section{The DexCanvas Dataset}
\label{sec:dataset}

DexCanvas combines 70 hours of real motion-capture demonstrations with 7{,}000 hours of physics-simulated rollouts that carry contact-force annotations. This section covers what we capture and how: the manipulation taxonomy and collection protocol (\S\ref{sec:taxonomy}), the capture system (\S\ref{sec:capture}), the conversion of raw markers into MANO trajectories (\S\ref{sec:mano}), and the released data (\S\ref{sec:release}). Section~\ref{sec:method} then describes how simulation adds forces and synthesizes new trajectories.

\begin{figure}[t]
    \centering
    \includegraphics[width=\linewidth]{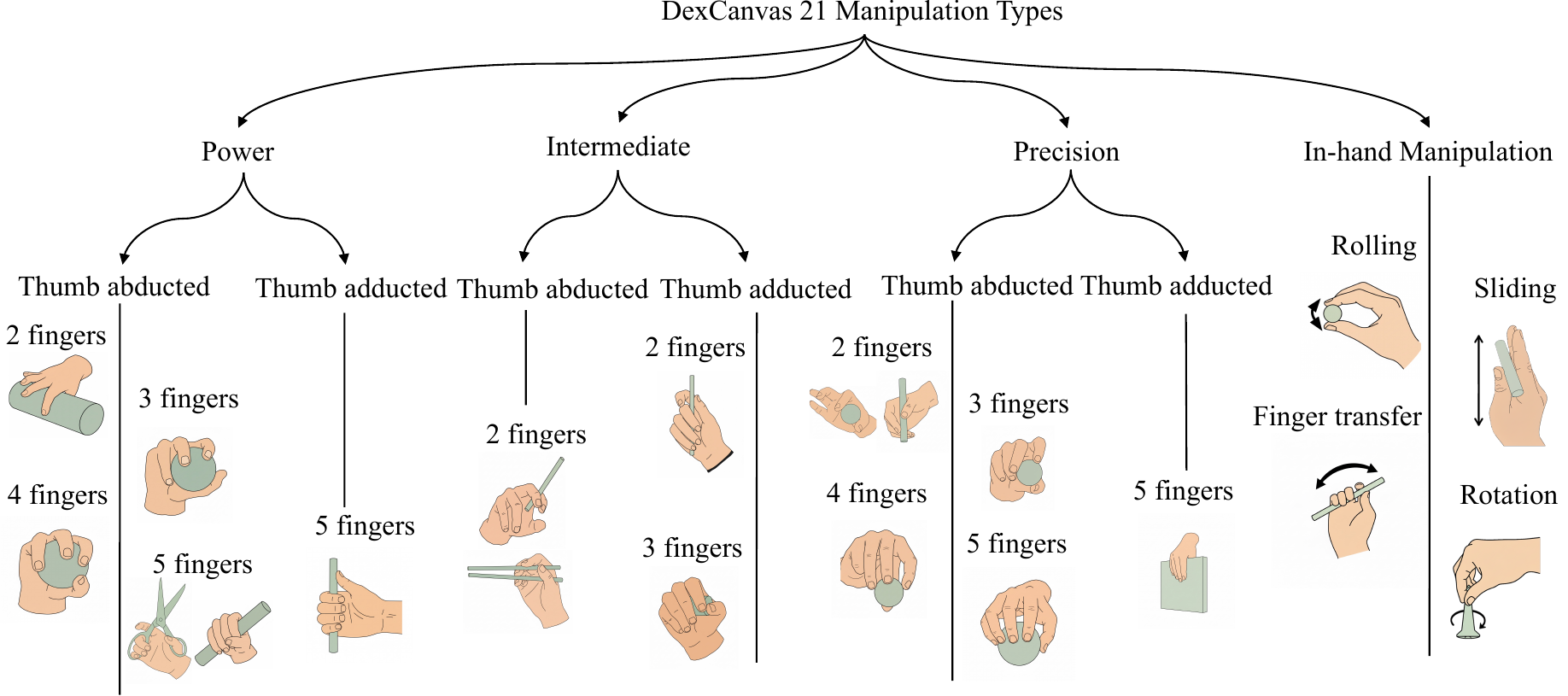}
    \caption{The 21 manipulation types in DexCanvas. Four categories organize the strategies: power (whole-hand grasps for stability), intermediate (combining power and precision), precision (fingertip control), and in-hand manipulation (dynamic reorientation). Grasp categories are subdivided by thumb position and the number of participating fingers.}
    \label{fig:manipulation_taxonomy}
\end{figure}

\subsection{Manipulation taxonomy and collection protocol}
\label{sec:taxonomy}

DexCanvas is organized around 21 manipulation types that draw their grasp types from the Cutkosky and GRASP taxonomies~\citep{feix2016grasp} and add in-hand manipulation, grouped into power grasps, intermediate grasps, precision grasps, and in-hand manipulations (rolling, sliding, finger transfer, and rotation). Within each grasp category, the types are distinguished by thumb position (abducted or adducted) and the number of participating fingers (Figure~\ref{fig:manipulation_taxonomy}). Collection follows this taxonomy, so the dataset spans the breadth of human dexterity by design.

\textbf{Objects.} The 29 objects (Table~\ref{tab:objects}) combine geometric primitives in multiple sizes to probe scale adaptation, weight-varied duplicates (suffix `H') with identical geometry to probe force modulation, and YCB objects~\citep{calli2017yale} such as a power drill and a pitcher that call for task-specific grasps.

\textbf{Protocol.} Five operators, trained on standardized reference demonstrations, performed 50 repetitions of each object--manipulation pair that the object's geometry admits. Every trial follows the same structure: object placement, manipulation, and return to a neutral pose. After removing trials with drops or extended occlusions, the real portion of DexCanvas contains 12{,}000 sequences totaling 70 hours.

\subsection{Multi-modal capture system}
\label{sec:capture}

\begin{figure}[t]
    \centering
    \includegraphics[width=0.82\linewidth]{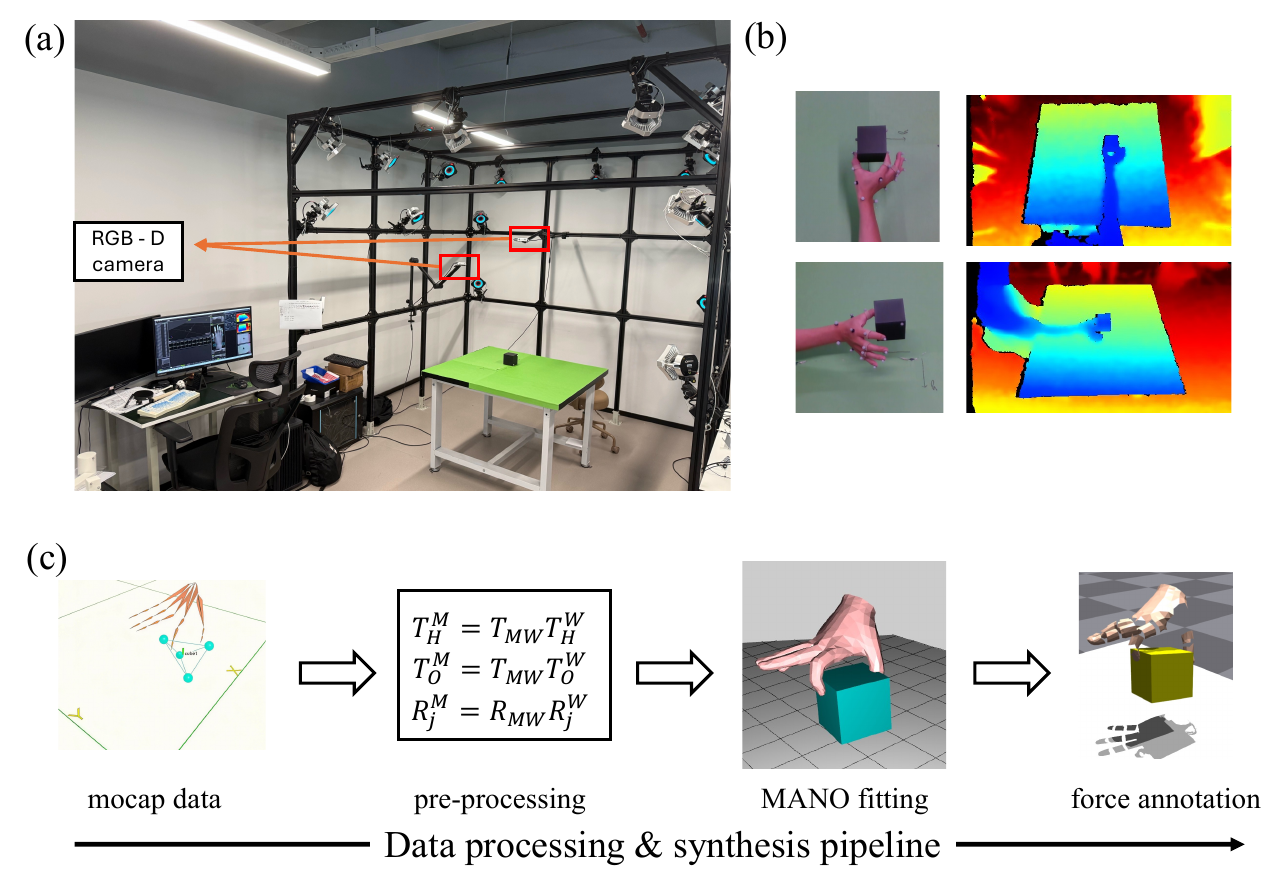}
    \caption{Capture setup and processing pipeline. (a) Room-scale optical motion capture with 22 infrared cameras and synchronized RGB-D sensors. (b) RGB and depth from the two calibrated RGB-D cameras, one camera per row. (c) Processing: raw markers are transformed into the MANO coordinate system, aligned with the RGB-D streams, and fitted with the MANO hand model; the resulting hand--object trajectories drive the RL reconstruction that adds force annotations.}
    \label{fig:capture}
\end{figure}

Twenty-two infrared cameras track reflective markers with millimeter precision, and two synchronized RGB-D cameras record front and side views (Figure~\ref{fig:capture}).

\textbf{Marker instrumentation.} Fourteen markers sit at anatomical landmarks on the right hand and four on each object. Every object is 3D-printed from its CAD model with the marker mounts built into the geometry, so the tracked pose refers to exactly the mesh and URDF used in simulation and no object-frame calibration is needed.

\textbf{Data streams.} Motion capture records the 6-DoF wrist pose, finger articulation, and object poses at 120\,Hz; the RGB-D cameras record at 30\,Hz. Brief marker occlusions are filled by interpolation, per-participant calibration accounts for hand size, and time synchronization gives frame-level correspondence across all streams.

\subsection{From markers to MANO trajectories}
\label{sec:mano}

We represent every demonstration with the MANO hand model~\citep{romero2022embodied}. For each participant, we estimate a fixed shape vector $\hat\beta \in \mathbb{R}^{10}$ by running the HaMeR hand-mesh regressor~\citep{pavlakos2024reconstructing} on RGB crops and robustly averaging its estimates. For every frame, we then solve for the wrist transform $T_t \in SE(3)$ and joint rotations $\theta_t$ that bring the virtual markers on the MANO surface onto the measured markers $\{x_{i,t}\}_{i=1}^{N}$:
\begin{equation}
\min_{T_t,\,\theta_t}\ \sum_{i=1}^{N} \big\| x_{i,t} - p_i(T_t, \hat\beta, \theta_t) \big\|_2^2 + \lambda_{\text{pose}}\,\psi(\theta_t),
\label{eq:mano_fit}
\end{equation}
where $p_i$ returns the $i$-th virtual marker and $\psi$ encodes soft joint-limit priors. Object poses come from rigid registration of the four object markers. A fixed transform maps hand and object poses from the motion-capture world frame into the MANO world frame, and the MANO forward model yields per-frame joints and meshes aligned with the RGB-D streams. Appendix~\ref{app:mano} gives the full procedure.

\subsection{Dataset contents and release}
\label{sec:release}

Every DexCanvas trajectory contains MANO shape and pose parameters, the wrist and object 6-DoF trajectories, 3D hand keypoints, and per-frame contacts: for each contacting hand link, the contact points and the normal forces it exerts on the object, expressed in the world, object, joint, and wrist frames. Contacts and forces are measured by the physics engine during the reconstruction rollouts of Section~\ref{sec:method}, and every trajectory has passed the physical-validity test defined there. Each trajectory links to the real capture that seeded it, whose camera streams provide the visual record of the demonstration. At the 120\,Hz capture rate, DexCanvas spans 3.0 billion frames.

The first public release on Hugging Face stores one complete trajectory per Parquet row: 111{,}857 force-annotated trajectories with 55.4 million frames at 200\,Hz (77 hours) and 96.5 million contact records, together with the complete dual-camera RGB streams of the 21{,}079 captures that seeded them, a URDF hand model for every hand shape, object meshes, and a standalone replay tool. Appendix~\ref{app:release} documents the schema and access.

\section{Real-to-Sim Force Reconstruction}
\label{sec:method}

\begin{figure}[t]
    \centering
    \includegraphics[width=0.85\linewidth]{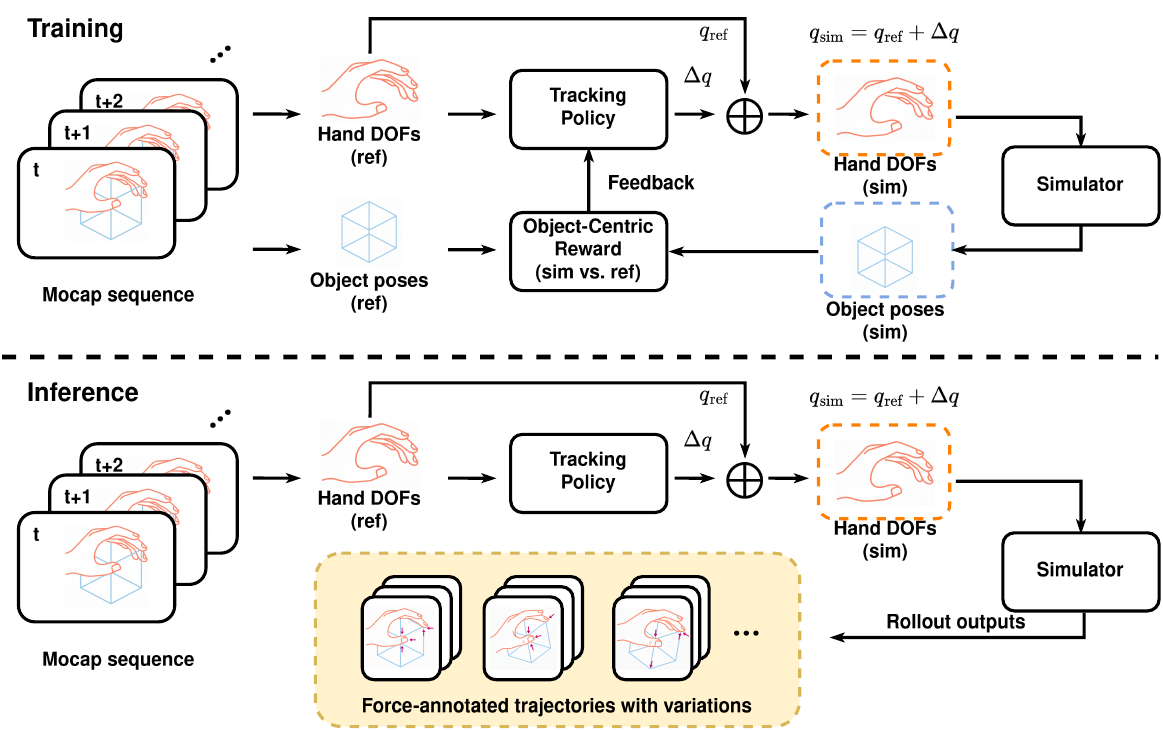}
    \caption{Real-to-sim reconstruction. Training (top): a tracking policy outputs residual corrections $\Delta q$ to the reference hand configuration $q_{\text{ref}}$ and is rewarded for driving the simulated object along its captured trajectory (the diagram shows the object-centric reward term). Inference (bottom): rollouts of the trained policy, from captured and perturbed initial conditions, yield force-annotated trajectories.}
    \label{fig:learning}
\end{figure}

Motion capture fixes geometry but not dynamics. Replaying even millimeter-accurate mocap open-loop in simulation fails: small systematic errors become interpenetrations or lost contacts, and the object drops or flies away. We close the loop with a tracking policy trained by reinforcement learning (Figure~\ref{fig:learning}). The policy acts as a residual controller that adds small corrections to the fitted MANO joint angles, so the simulated hand keeps the human gesture while the corrections establish stable contact. The policy adjusts only kinematics; the physics simulator measures the resulting forces as the policy executes. Reconstruction thereby becomes a tracking-control problem in which the simulator enforces physical consistency.

\subsection{One policy for all object--manipulation pairs}

A single policy $\pi_\theta$ serves every object--manipulation pair, with object geometry and manipulation type supplied as conditioning. Each demonstration defines an episode in IsaacGym~\citep{makoviychuk2021isaac}: the simulated MANO hand, a floating-base system with actuated fingers, starts in the captured configuration and the object at its captured pose.

\textbf{Observation.} At each step the policy observes hand proprioception (wrist state, finger joints, fingertip and keypoint positions relative to the object), the object pose, the reference object pose from motion capture together with a short-horizon preview of its future position, the accumulated residuals, a PointNet encoding of the object point cloud, a compact geometry descriptor, the manipulation type, and contact feedback: per-keypoint force magnitudes and directions alongside the contacts expected from the demonstration.

\textbf{Action.} The policy outputs increments for wrist translation, wrist rotation, and finger joints, which are accumulated with exponential decay into the executed residual
\begin{equation}
u_t = \tau\, u_{t-1} + a_t, \qquad \tau \in [0.8, 0.95],
\label{eq:action_filter}
\end{equation}
and added to the fitted MANO configuration. The accumulation smooths the commanded motion, which keeps contacts from breaking under abrupt wrist or finger changes.

\textbf{Reward and termination.} The reward combines object position and rotation tracking (following~\citep{objdex2024}), coverage of the expected contacts during the contact phase, object stability after it, a small survival bonus, an L1 penalty on the accumulated residuals that keeps the hand close to the human gesture, and a penalty for deviating from the reference. An episode terminates when the trajectory completes or when the object deviates from its reference by more than 5\,cm. We train with PPO~\citep{schulman2017proximal}. Appendix~\ref{app:policy} lists the observation, action, and reward components in full.

\subsection{Force extraction and data synthesis}

A rollout is \emph{physically valid} when it completes the full demonstration without termination, that is, with the simulated object never more than 5\,cm from its captured trajectory. From every valid rollout we record per-frame contact points and contact forces directly from the physics engine. These are forces with which a MANO hand, under simulated physics, reproduces the demonstrated object motion.

The trained policy also generates new data. We roll it out from initial object poses shifted in the horizontal plane by up to 20\% of the object's length and width and keep only the physically valid rollouts. Each real demonstration thereby seeds many distinct physically valid variants, which expands the 70 captured hours into 7{,}000 hours of force-annotated trajectories.

\section{Experiments}
\label{sec:experiments}

We evaluate DexCanvas along five questions. (1)~How reliably does one policy reconstruct the captured demonstrations, and how well does it synthesize new ones? (2)~Does the reconstruction transfer to objects outside the capture set? (3)~What do the contact-force annotations look like? (4)~Do these annotations improve downstream policy learning? (5)~Do the trajectories transfer to robot hands beyond the MANO embodiment?

\subsection{Reconstruction and synthesis}
\label{sec:reconstruction}

We evaluate the unified policy on 61 object--manipulation pairs over nine geometric primitives, once from the captured initial conditions (nominal) and once with the initial object pose shifted in the horizontal plane by up to 20\% of the object's length and width (perturbed). Success is physical validity as defined above.

\begin{figure}[t]
\centering
\begin{tikzpicture}
  \node[inner sep=0pt, anchor=south west] (hist) at (0,0) {\includegraphics[width=\linewidth]{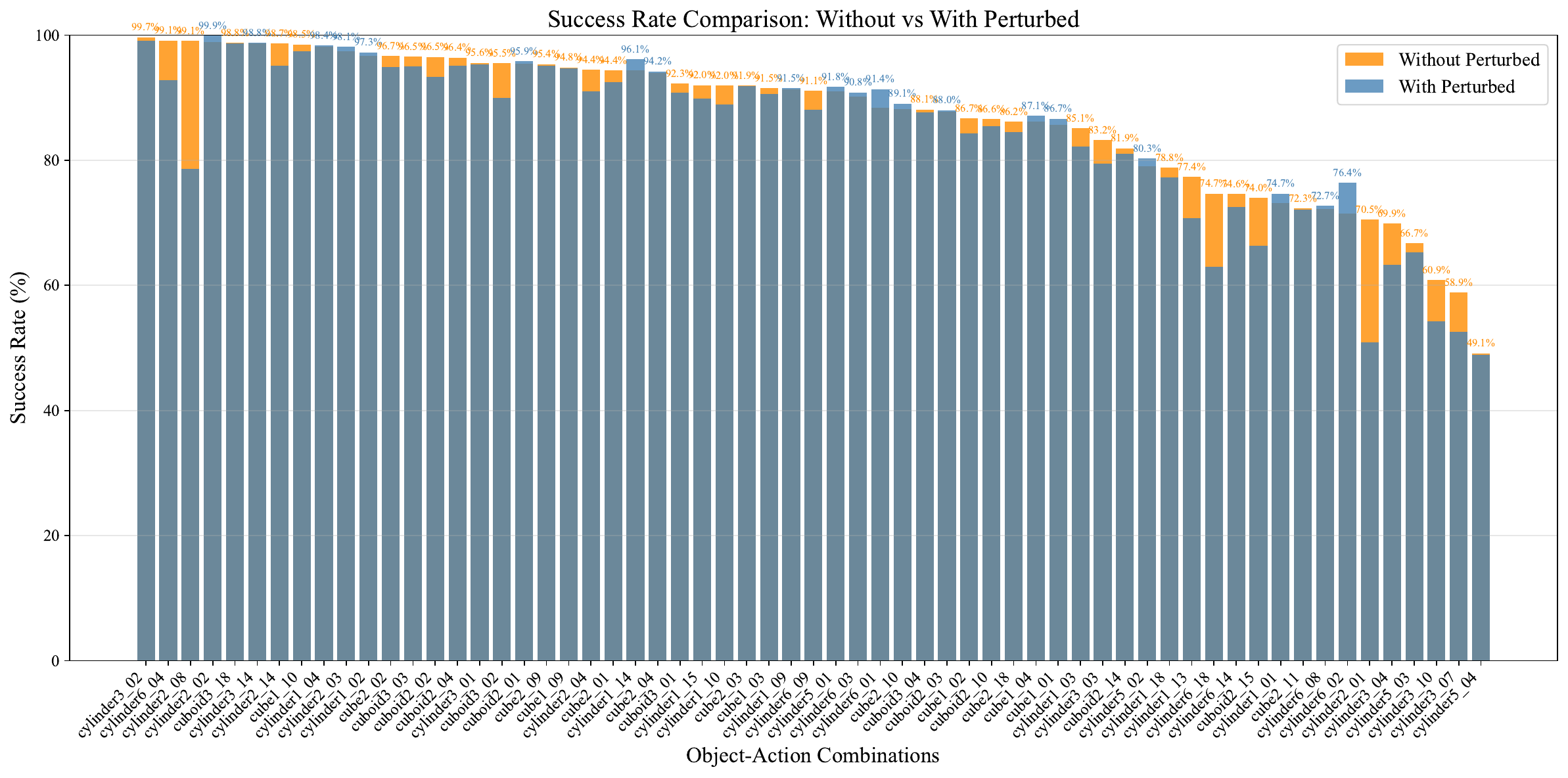}};
  \begin{scope}[x={(hist.south east)}, y={(hist.north west)}]
    \fill[white] (0.345,0.955) rectangle (0.695,0.995);
  \end{scope}
\end{tikzpicture}
\caption{Success rate of the unified policy on 61 object--manipulation pairs, labelled \texttt{object\_ID} with the manipulation-type code of the dataset's \texttt{action\_label} field. For each pair the nominal (orange, ``without perturbed'') and perturbed (blue, ``with perturbed'') success rates are drawn as overlaid bars; the visible colored segment marks the condition with the higher rate. Perturbation shifts the initial object pose by up to 20\% of the object's length and width.}
\label{fig:success_hist}
\end{figure}

The policy reproduces \textbf{87.54\%} of demonstrations under nominal conditions (Figure~\ref{fig:success_hist}). Under perturbation, success falls by only 6.68 percentage points to \textbf{80.86\%}, so most perturbed rollouts are themselves physically valid trajectories. Retaining these rollouts is what turns 70 hours of demonstrations into 7{,}000 hours of data. Nominal success exceeds 90\% for about half of the pairs and falls below 70\% only for the mini cylinder \texttt{cylinder5} ($D2\times2.7$\,cm) and the long, slender \texttt{cylinder3} ($D3\times10$\,cm), where small positional errors suffice to break fingertip contact.

Perturbation can also raise success. Several pairs succeed more often from perturbed starts than from the captured one, among them \texttt{cylinder6\_02}, \texttt{cylinder6\_01}, and \texttt{cylinder1\_14}. Elongated cylinders tip easily, and some human demonstrations approach them in a tightly constrained configuration with little clearance; a small shift of the initial pose moves the interaction into a region with more favorable contact timing and fewer early failures. Synthesis therefore explores physically valid neighbors of each demonstration.

\subsection{Generalization to unseen objects}
\label{sec:unseen}

\begin{figure}[t]
    \centering
    \includegraphics[width=0.9\linewidth]{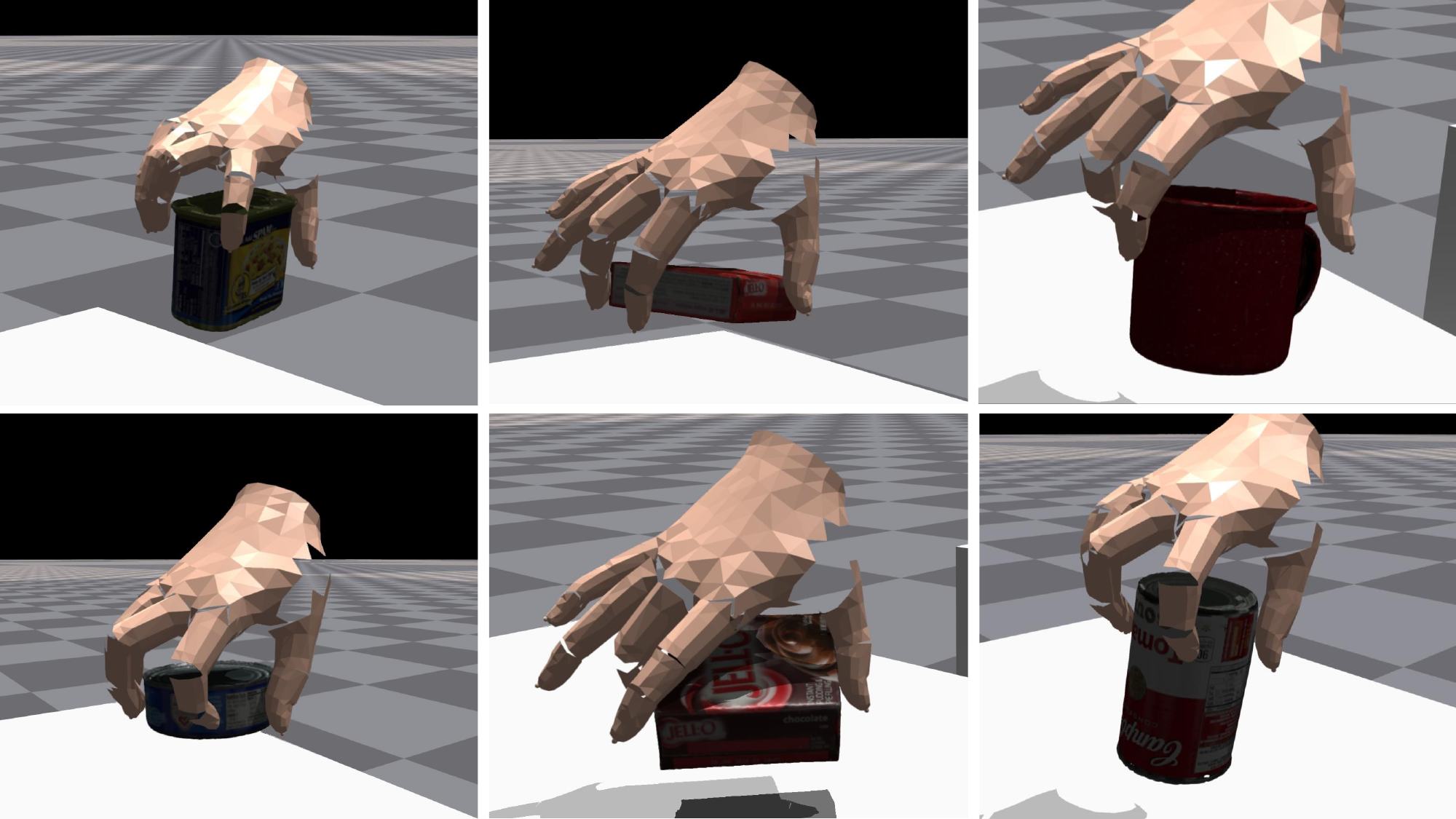}
    \caption{The unified policy manipulating six YCB objects that were never captured. Top row, left to right: potted meat can, pudding box, mug. Bottom row: tuna fish can, gelatin box, tomato soup can. Table~\ref{tab:unseen} reports per-object success.}
    \label{fig:unseen_objects}
\end{figure}

We apply the unified policy, unchanged, to six objects from the YCB set~\citep{calli2017yale} that were never captured: cylindrical cans, rectangular boxes, and a handled mug (Figure~\ref{fig:unseen_objects}). Each object is evaluated in about 1{,}290 rollouts with the same success criterion.

\begin{table}[t]
\centering
\small
\caption{Success of the unified policy on six YCB objects absent from the capture set.}
\label{tab:unseen}
\begin{tabular}{@{}lrrr@{}}
\toprule
\textbf{YCB object} & \textbf{Successes} & \textbf{Rollouts} & \textbf{Success rate} \\
\midrule
005 tomato soup can & 599 & 1{,}289 & 46.5\% \\
007 tuna fish can & 1{,}141 & 1{,}287 & 88.7\% \\
008 pudding box & 818 & 1{,}287 & 63.6\% \\
009 gelatin box & 1{,}081 & 1{,}295 & 83.5\% \\
010 potted meat can & 840 & 1{,}238 & 67.8\% \\
025 mug & 1{,}088 & 1{,}287 & 84.6\% \\
\midrule
\textbf{Pooled} & \textbf{5{,}567} & \textbf{7{,}683} & \textbf{72.5\%} \\
\bottomrule
\end{tabular}
\end{table}

Pooled over all rollouts, the policy succeeds \textbf{72.5\%} of the time on objects it has never seen (Table~\ref{tab:unseen}). The squat tuna can, the gelatin box, and the mug reach 84--89\%, close to in-distribution performance. The tall tomato soup can is hardest at 46.5\%, mirroring the tipping behavior of elongated cylinders in Section~\ref{sec:reconstruction}. The policy that annotates and expands DexCanvas thus transfers to new object instances without retraining.

\subsection{Contact-force annotations}
\label{sec:force_quality}

We examine the annotations at two levels: the force profile of a single trajectory, and the contact distribution of each manipulation type.

\begin{figure}[t]
    \centering
    \includegraphics[width=\linewidth]{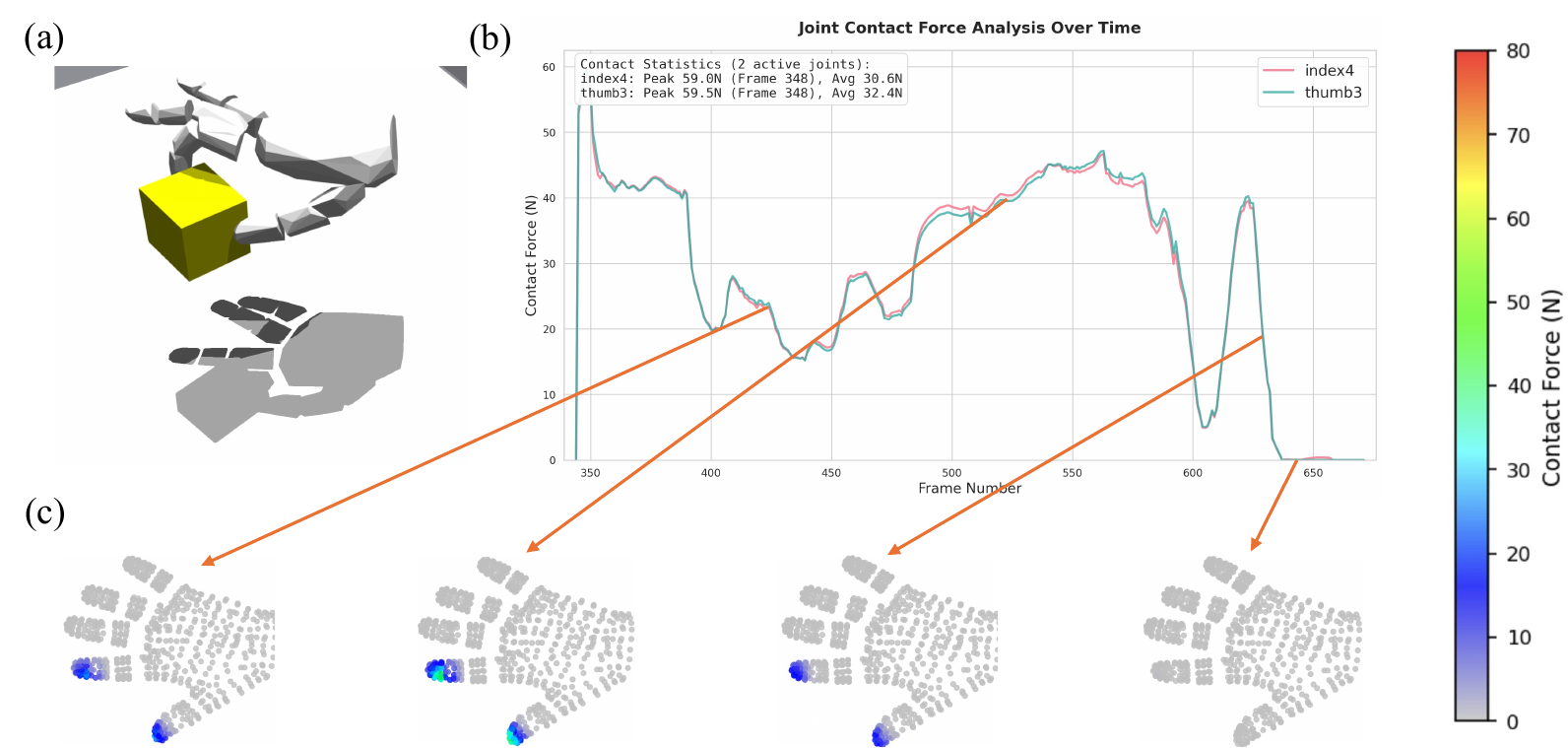}
    \caption{Contact forces along one reconstructed trajectory. (a) The manipulation reproduced in simulation. (b) Contact force over time on the two active links, the distal segments of the index finger (\texttt{index4}) and thumb (\texttt{thumb3}), with peak and mean statistics. (c) Force distribution rendered on the hand at four keyframes marked in (b).}
    \label{fig:force_viz}
\end{figure}

\textbf{Trajectory level.} Figure~\ref{fig:force_viz} follows one manipulation. The force profiles in (b) rise sharply at contact onset, vary with the object's motion, and fall to zero at release. The two opposing digits, index finger and thumb, press with nearly identical normal force throughout, as the balance of the squeeze across the object requires. The renderings in (c) place the forces on the hand regions that touch the object. Every frame of every trajectory carries this per-link force record.

\begin{figure}[t]
    \centering
    \includegraphics[width=0.9\linewidth]{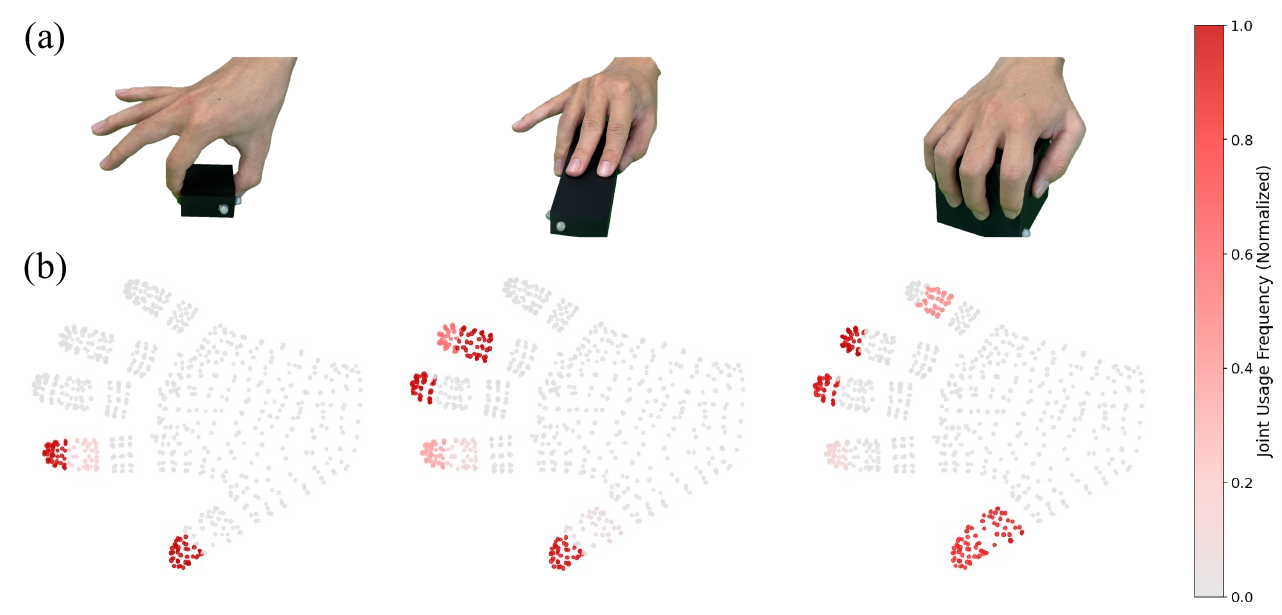}
    \caption{Contact signatures of manipulation types. (a) Three manipulation types from the taxonomy. (b) Contact frequency of each hand region, aggregated over 100 simulated rollouts of each type.}
    \label{fig:contact_dist}
\end{figure}

\textbf{Manipulation level.} For three manipulation types, we aggregate contact statistics over 100 rollouts each (Figure~\ref{fig:contact_dist}). Each type has a distinct signature: in the two-finger pinch on the left, contact concentrates on the thumb and index fingertips, while grasps that recruit more fingers spread contact across additional fingertips and along the thumb. Manipulation types thus differ measurably in where the hand touches the object, which gives manipulation-specific signal for learning contact-aware control.

\subsection{Downstream policy learning}
\label{sec:dp_utility}

Do the contact channels of DexCanvas make better policies? We train Diffusion Policy~\citep{chi2023diffusion} with FiLM conditioning~\citep{perez2018film} on object identity and manipulation type to control the simulated MANO hand and reproduce DexCanvas trajectories. Training uses a window of each trajectory from 50 frames before the object first moves to 50 frames after it comes to rest. In the \emph{with-contact} condition, the policy additionally observes per-step contact-force vectors and magnitudes, both in training and at execution; in the \emph{no-contact} condition these features are removed and the policy observes kinematic state only. Both conditions share architecture, data, and hyperparameters (Appendix~\ref{app:diffusion}). A trial is one policy execution on one trajectory; it succeeds when the object's positional error relative to the reference trajectory stays below 3\,cm and the cumulative rotation error below $30^{\circ}$ across the three axes throughout, a stricter criterion than the 5\,cm used for reconstruction.

\begin{table}[t]
\centering
\small
\caption{Diffusion Policy success with and without DexCanvas contact-force features. Manipulation IDs are the manipulation-type codes of Figure~\ref{fig:success_hist}; cylinder1 and cube1 are geometric primitives (Table~\ref{tab:objects}).}
\label{tab:dp_ablation}
\begin{tabular}{@{}llrrrr@{}}
\toprule
\textbf{Object} & \textbf{Manip.\ ID} & \textbf{Trials} & \textbf{With contact} & \textbf{No contact} & \textbf{$\Delta$ (pp)} \\
\midrule
cylinder1 & 01 & 550 & 81.6\% & 71.9\% & $+9.7$ \\
cylinder1 & 02 & 550 & 82.4\% & 68.4\% & $+14.0$ \\
cylinder1 & 03 & 550 & 61.3\% & 56.1\% & $+5.2$ \\
cylinder1 & 04 & 550 & 32.2\% & 38.6\% & $-6.4$ \\
cylinder1 & 09 & 550 & 99.5\% & 88.9\% & $+10.6$ \\
\midrule
cube1 & 01 & 3{,}943 & 95.6\% & 80.5\% & $+15.1$ \\
cube1 & 02 & 3{,}943 & 99.2\% & 78.5\% & $+20.7$ \\
cube1 & 03 & 3{,}943 & 72.1\% & 64.7\% & $+7.4$ \\
cube1 & 04 & 3{,}943 & 69.3\% & 48.3\% & $+21.0$ \\
\midrule
\textbf{Pooled} & --- & \textbf{18{,}522} & \textbf{82.2\%} & \textbf{67.5\%} & $\mathbf{+14.7}$ \\
\bottomrule
\end{tabular}
\end{table}

Observing contact features raises pooled success from \textbf{67.5\%} to \textbf{82.2\%}, a gain of \textbf{14.7 percentage points} over 18{,}522 trials (Table~\ref{tab:dp_ablation}). The gain holds in eight of nine object--manipulation cells and reaches 20.7 and 21.0 points on two cube manipulations. DexCanvas thus supports training contact-aware policies, and the force channels translate directly into higher success.

\subsection{Cross-embodiment transfer}
\label{sec:cross_embodiment}

\begin{figure}[t]
  \centering
  \begin{tikzpicture}[
      panel/.style={inner sep=0pt},
      lbl/.style={font=\sffamily\footnotesize, text=DexInk, anchor=north, inner sep=0pt},
      tag/.style={font=\sffamily\bfseries\footnotesize, text=DexInk, anchor=north west, inner sep=0pt}]
    \node[panel] (mano) at (0,0) {\includegraphics[height=3.8cm]{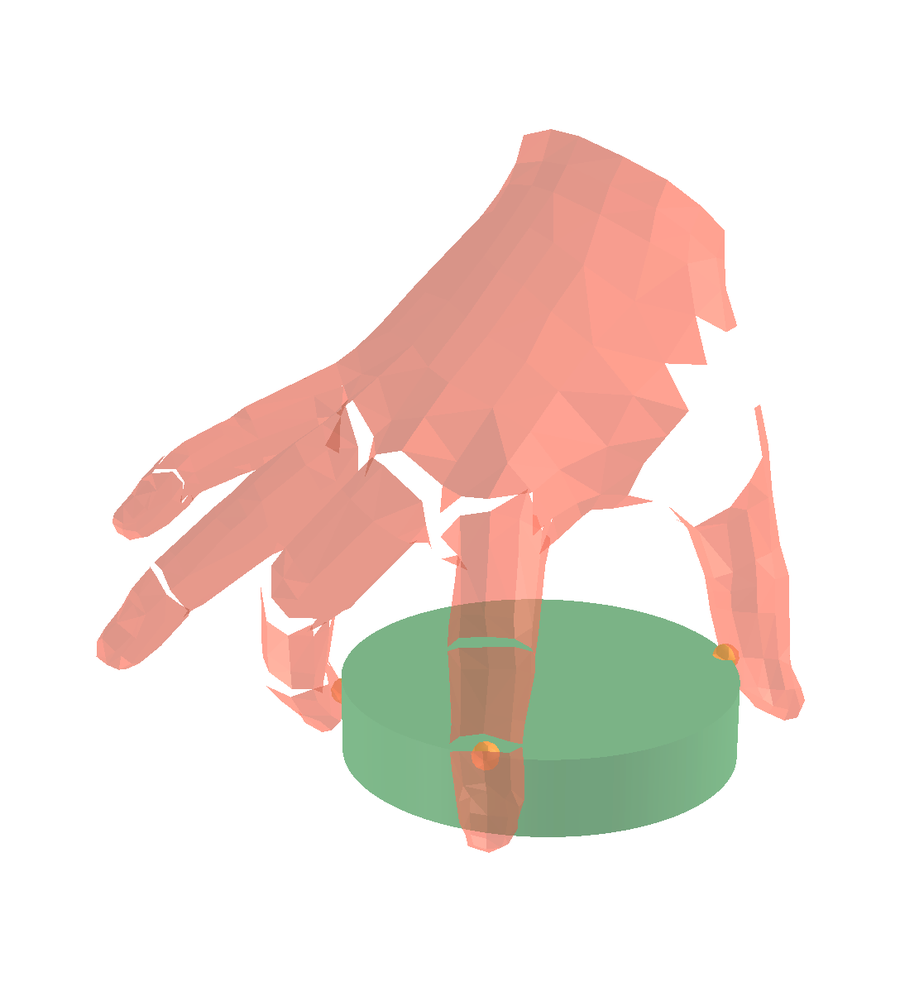}};
    \node[panel, right=1.55cm of mano] (d021) {\includegraphics[height=3.8cm]{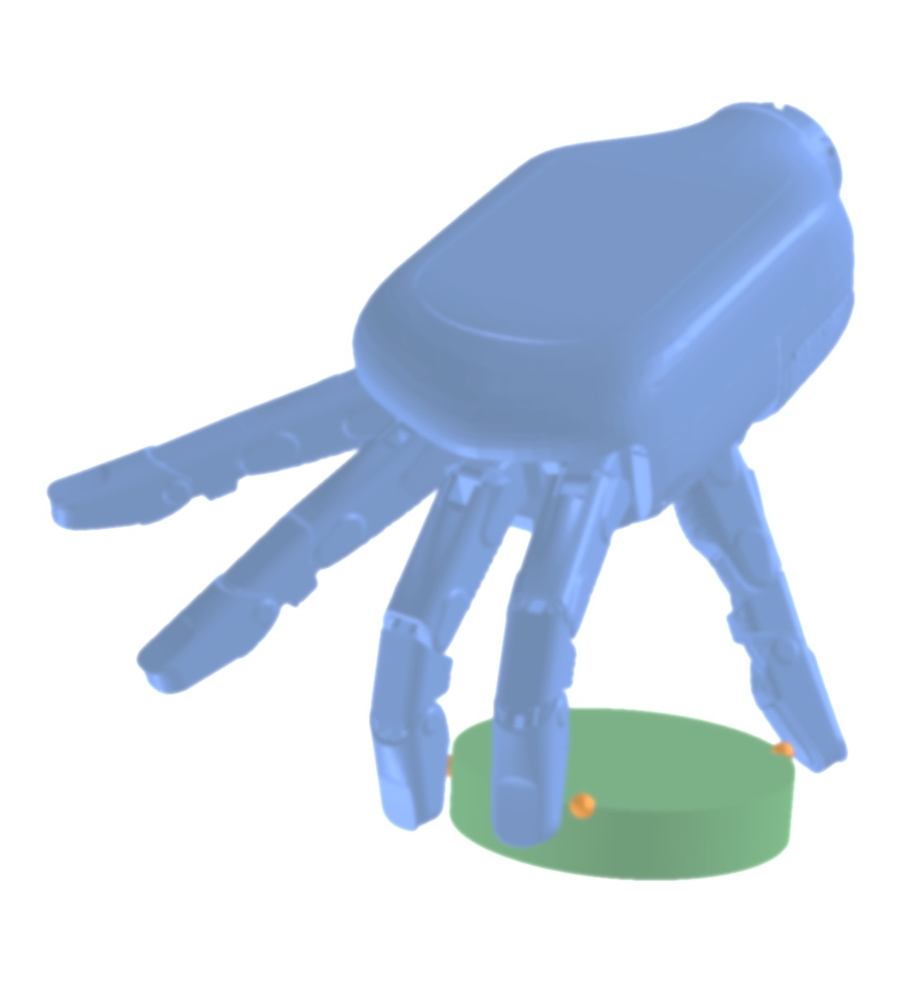}};
    \node[panel, right=0.25cm of d021] (d021p) {\includegraphics[height=3.8cm]{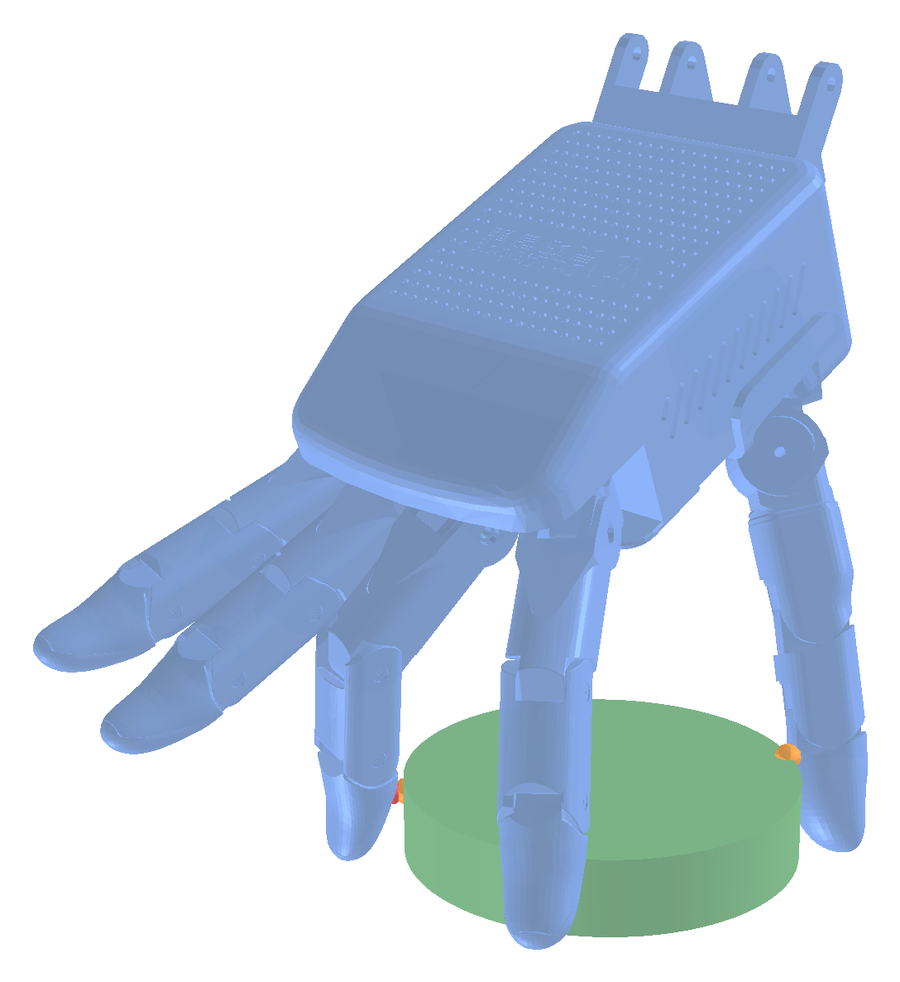}};
    \draw[-{Stealth[length=5pt,width=4pt]}, line width=0.7pt, DexInkSixty]
      ($(mano.east)+(0.12,0)$) -- ($(d021.west)+(-0.12,0)$)
      node[midway, above=3pt, font=\sffamily\scriptsize, text=DexInkSixty, align=center] {contact-driven\\retargeting};
    \node[lbl] at ($(mano.south)+(0,-0.12)$) {Human demonstration (MANO)};
    \node[lbl] at ($(d021.south)+(0,-0.12)$) {DexHand021};
    \node[lbl] at ($(d021p.south)+(0,-0.12)$) {DexHand021Pro};
    \node[tag] at ($(mano.north west)+(0,0.05)$) {(a)};
  \end{tikzpicture}

  \vspace{10pt}
  \begin{tikzpicture}
    \node[inner sep=0pt] (rl) {\includegraphics[width=\linewidth]{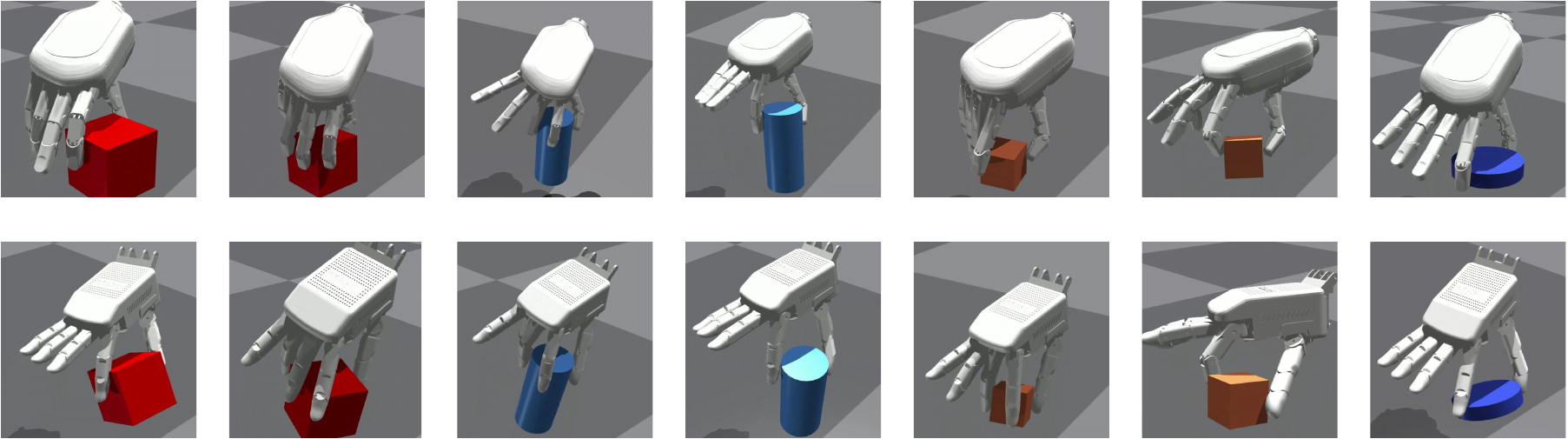}};
    \node[font=\sffamily\bfseries\footnotesize, text=DexInk, anchor=south west, inner sep=0pt] at ($(rl.north west)+(0,0.08)$) {(b)};
  \end{tikzpicture}
  \caption{Cross-embodiment transfer. (a) Contact-driven retargeting maps the contacts of a human demonstration (orange points on the object) to the corresponding links of DexHand021 and DexHand021Pro, which differ in kinematic structure. (b) RL-reconstructed rollouts of retargeted trajectories on DexHand021 (top row) and DexHand021Pro (bottom row) for four objects: red cube (columns 1--2), blue cylinder (columns 3--4), brown cube (columns 5--6), and blue disk (column 7); paired columns show the same grasp from two viewpoints.}
  \label{fig:cross_embodiment}
\end{figure}

DexCanvas describes manipulation with a human hand, and it transfers to robot hands through contact. We retarget MANO demonstrations to two dexterous robot hands with different kinematic structures: DexHand021~\citep{dexhand021_website} and DexHand021Pro~\citep{dexhand021pro_website} (Figure~\ref{fig:cross_embodiment}a). For each frame, we extract contact tokens from the MANO hand--object interaction, map each to the corresponding finger link of the robot hand, and express its target contact point in the object frame. We then solve for the robot's wrist pose and finger joints that bring its contact links onto these targets:
\begin{equation}
E = w_{\text{target}}E_{\text{target}} + w_{\text{pen}}E_{\text{pen}} + w_{\text{self}}E_{\text{self}} + w_{\text{joint}}E_{\text{joint}} + w_{\text{temp}}E_{\text{temp}},
\label{eq:retarget}
\end{equation}
where $E_{\text{target}}$ preserves the demonstrated contacts and the remaining terms penalize hand--object penetration, self-collision, joint-limit violation, and temporal discontinuity (Appendix~\ref{app:retargeting}). Matching contacts in the object frame, instead of matching joint angles, lets one human demonstration drive hands with different kinematics.

Retargeted kinematics still needs physical repair, and the RL reconstruction stage of Section~\ref{sec:method} supplies it unchanged. Applied to the retargeted trajectories, it produces stable rollouts on both hands across cubes, cylinders, and disks (Figure~\ref{fig:cross_embodiment}b), with motion patterns that follow each hand's kinematic structure. Contact-anchored retargeting combined with residual RL has since been scaled to four robot hands and to bimanual execution on physical hardware~\citep{yang2026contact}.

\section{Discussion and Outlook}

DexCanvas turns human demonstrations into physically grounded training data for dexterous robots: real manipulation strategies, systematic taxonomy coverage, and per-finger contact forces at the scale of thousands of hours. Its central mechanism, a single residual policy that reconstructs demonstrations in simulation, annotates forces, synthesizes new trajectories, generalizes to unseen objects, and executes retargeted motion on robot hands.

The 21 manipulation types are building blocks that compose into free-form, multi-stage behaviors on everyday objects. The camera streams open visuomotor policy learning and world-model-based perception synthesis on top of the force-annotated trajectories. Physical grounding also makes language measurable: descriptors such as ``gentle'' and ``firm'' contact acquire quantitative anchors in force, which connects DexCanvas to vision-language-action pretraining. We release DexCanvas, with its hand and object assets and replay tooling, to accelerate research on contact-aware reward shaping, exploration, and imitation for dexterous manipulation.

\bibliographystyle{abbrvnat}
\bibliography{dexcanvas}

\clearpage
\appendix
\section{Data Release and Access}
\label{app:release}

DexCanvas is hosted at \url{https://huggingface.co/datasets/Dex-GEM/DexCanvas} under the Open Database License (ODbL~v1.0). The release contains trajectory Parquet shards, the dual-camera JPEG streams of every source capture packaged as WebDataset-compatible archives, capture and trajectory-to-capture indexes, kinematic assets, a replay tool, and checksums. Table~\ref{tab:release} summarizes its contents and Table~\ref{tab:data_structure} the trajectory schema.

\begin{table}[H]
\centering
\small
\caption{Contents of the public DexCanvas release.}
\label{tab:release}
\begin{tabular}{@{}lr@{}}
\toprule
\textbf{Item} & \textbf{Count} \\
\midrule
Trajectories (one Parquet row each) & 111{,}857 \\
Frames & 55{,}442{,}683 \\
Contact records & 96{,}484{,}004 \\
Action labels (manipulation-type codes) & 32 \\
Object categories & 22 \\
Right-hand MANO shape identities (URDF bundles) & 8 \\
Dual-camera source captures & 21{,}079 \\
Camera images (JPEG) & 22{,}212{,}754 \\
\bottomrule
\end{tabular}
\end{table}

\begin{table}[H]
\centering
\small
\caption{Trajectory schema. $T$ is the number of frames; positions are in meters, rotations are axis-angle vectors in radians, and forces are in newtons.}
\label{tab:data_structure}
\begin{tabularx}{\linewidth}{@{}l>{\ttfamily\raggedright\arraybackslash}Xl@{}}
\toprule
\textbf{Group} & \textbf{\normalfont\bfseries Field} & \textbf{Shape} \\
\midrule
Trajectory & trajectory\_id, action\_label, frame\_rate\_hz, num\_frames & scalars \\
 & timestamps\_s & $[T]$ \\
\midrule
Hand & mano\_shape & $[10]$ \\
 & global\_position\_world\_m, global\_rotation\_world\_rotvec\_rad & $[T,3]$ \\
 & mano\_pose\_rotvec\_rad & $[T,48]$ \\
 & joint\_positions\_world\_m (16 keypoints + 5 fingertips) & $[T,21,3]$ \\
 & urdf\_dof, urdf\_dof\_target (6 floating-base + 22 finger DoF) & $[T,28]$ \\
\midrule
Object & category & scalar \\
 & position\_world\_m, rotation\_world\_rotvec\_rad & $[T,3]$ \\
\midrule
Contacts & {\normalfont per frame, a list over contacting hand links with} joint\_name{\normalfont, the total force in world, joint, object, and wrist frames, and contact points with normal forces} & $[T][\cdot]$ \\
\bottomrule
\end{tabularx}
\end{table}

Because a row holds a complete trajectory, streaming is the recommended access pattern:

\begin{lstlisting}[language=Python]
from datasets import load_dataset

trajectories = load_dataset("Dex-GEM/DexCanvas", split="train", streaming=True)

for trajectory in trajectories:
    hand = trajectory["hands"][0]
    pose = hand["mano_pose_rotvec_rad"]          # [T, 48]
    obj = trajectory["objects"][0]["position_world_m"]  # [T, 3]
    contacts = trajectory["contacts"]            # per-frame contact lists
    break
\end{lstlisting}

Each trajectory references the real capture that seeded it through \nolinkurl{metadata/trajectory_image_links.parquet}, so its dual-camera stream can be retrieved from the image archives. The included Rerun player reconstructs the hand through URDF forward kinematics and displays the object mesh, contact points, and scaled contact-force arrows from the Parquet data alone:

\begin{lstlisting}[language=bash]
python replay/replay_rerun.py --data data --assets assets \
    --trajectory-id traj_00000001 --spawn
\end{lstlisting}

Trajectories can be filtered by manipulation type (\texttt{action\_label}) to study a specific strategy and by object category to compare geometry.

\section{Objects}
\label{app:objects}

\begin{table}[H]
\centering
\scriptsize
\setlength{\tabcolsep}{4pt}
\caption{Objects in DexCanvas with dimensions, weights, placement, and renderings. Suffix `H' marks a heavier duplicate with identical geometry.}
\label{tab:objects}
\begin{tabularx}{\linewidth}{@{}l l l l X c@{}}
\toprule
\textbf{ID} & \textbf{Type} & \textbf{Size (cm)} & \textbf{Mass (g)} & \textbf{Placement} & \textbf{Rendering} \\
\midrule
cube1       & Large cube          & $8 \times 8 \times 8$   & 125 & --                              & \raisebox{-0.3\height}{\includegraphics[width=0.6cm]{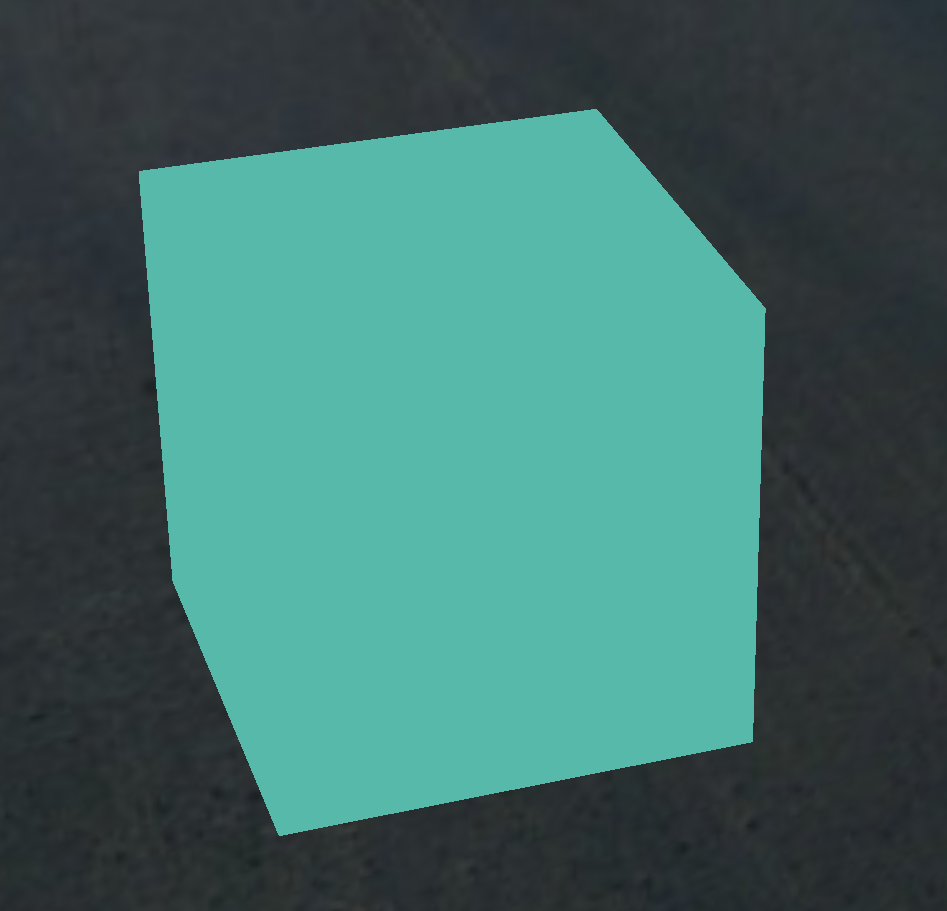}} \\
cube2       & Small cube          & $5 \times 5 \times 5$   & 37  & --                              & \raisebox{-0.3\height}{\includegraphics[width=0.6cm]{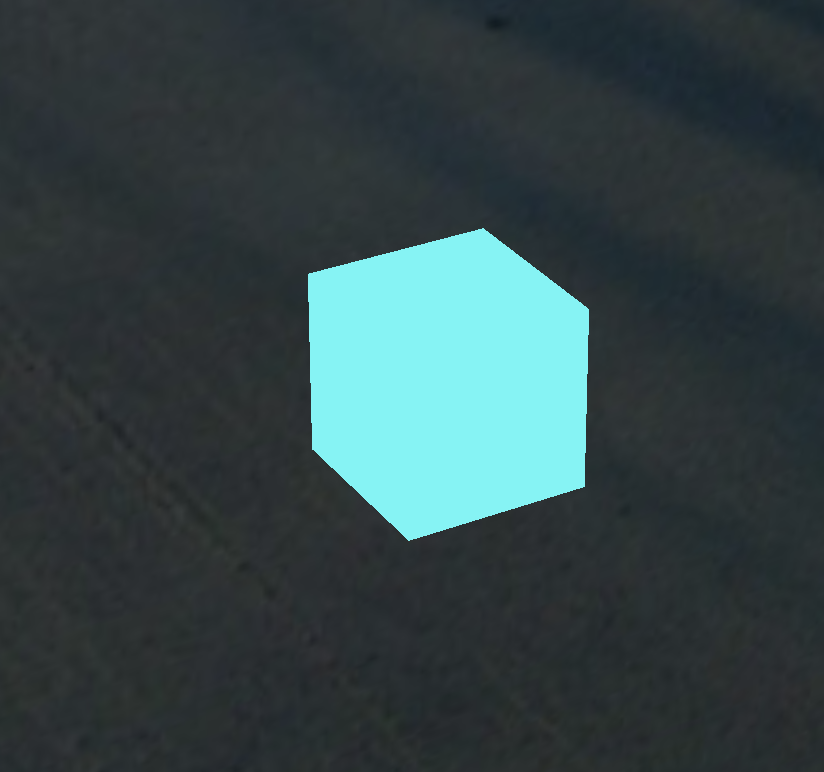}} \\
cuboid1     & Flat cuboid         & $15 \times 15 \times 2$ & 128 & Large face down; side face down & \raisebox{-0.3\height}{\includegraphics[width=0.6cm]{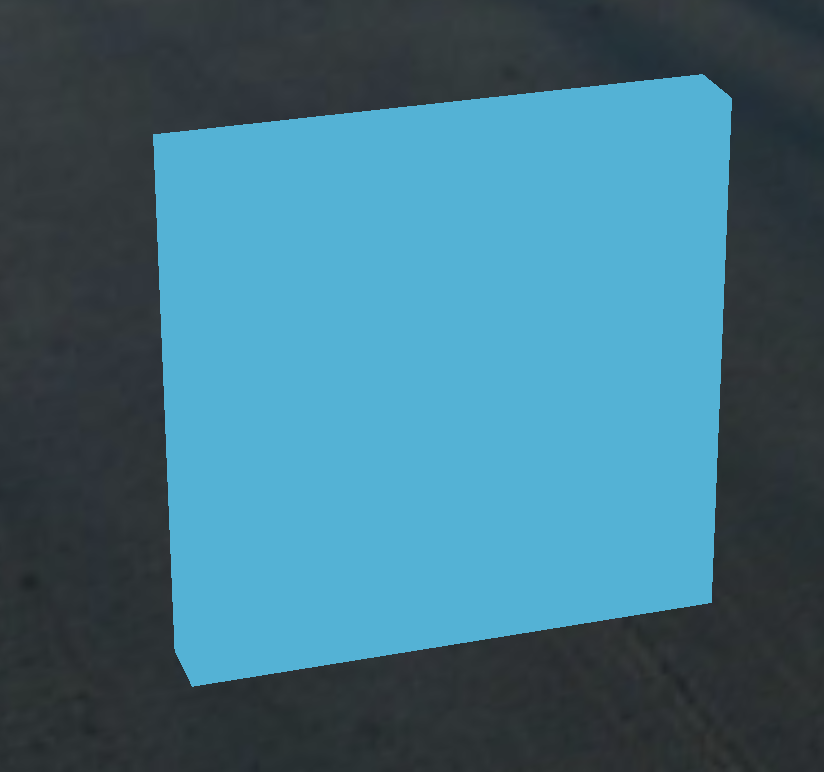}} \\
cuboid2     & Long cuboid         & $15 \times 5 \times 2.5$& 55  & Long edge along $x$-axis        & \raisebox{-0.3\height}{\includegraphics[width=0.6cm]{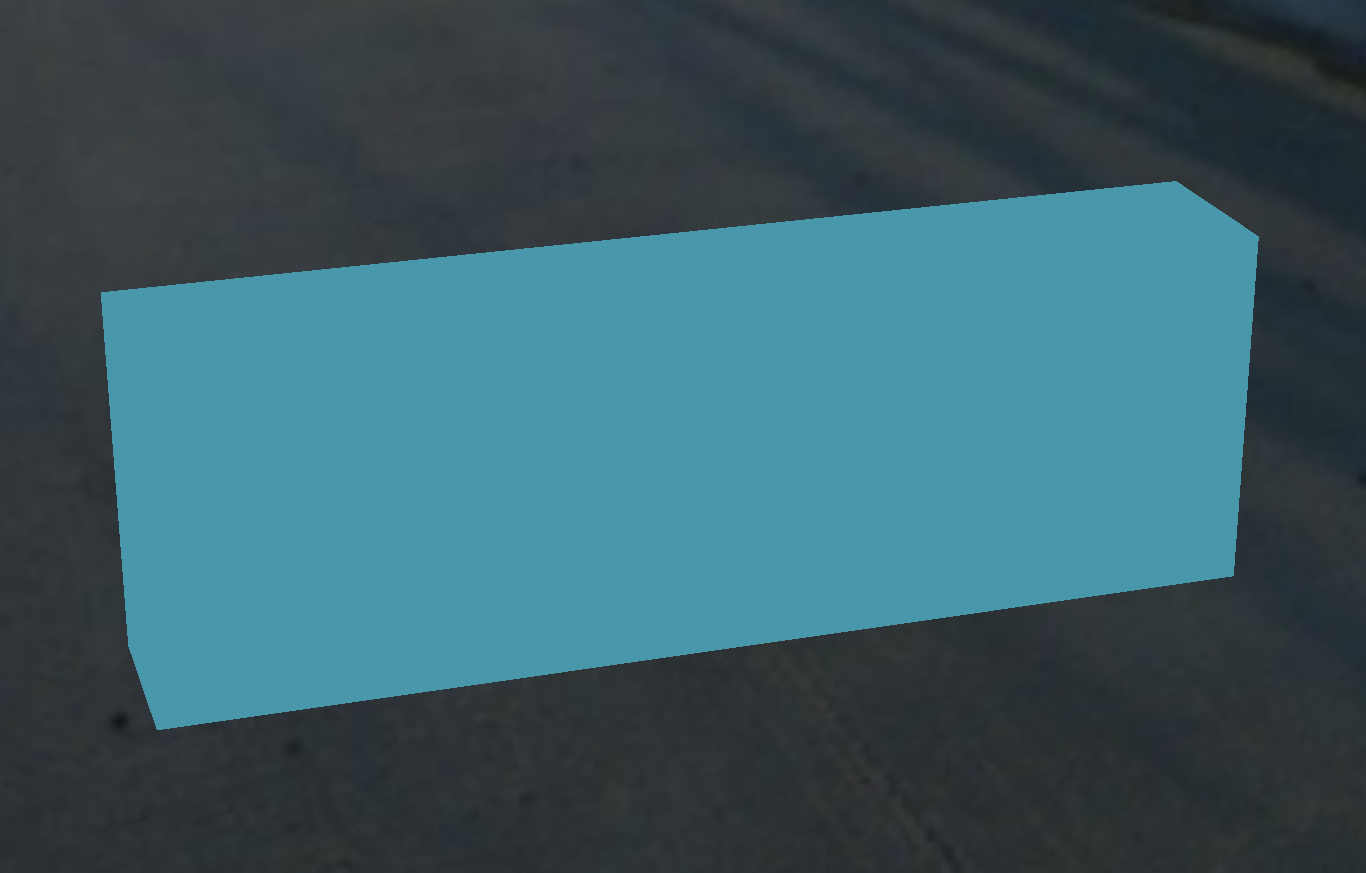}} \\
cuboid3     & Small cuboid        & $5 \times 5 \times 2.5$ & 22  & --                              & \raisebox{-0.3\height}{\includegraphics[width=0.6cm]{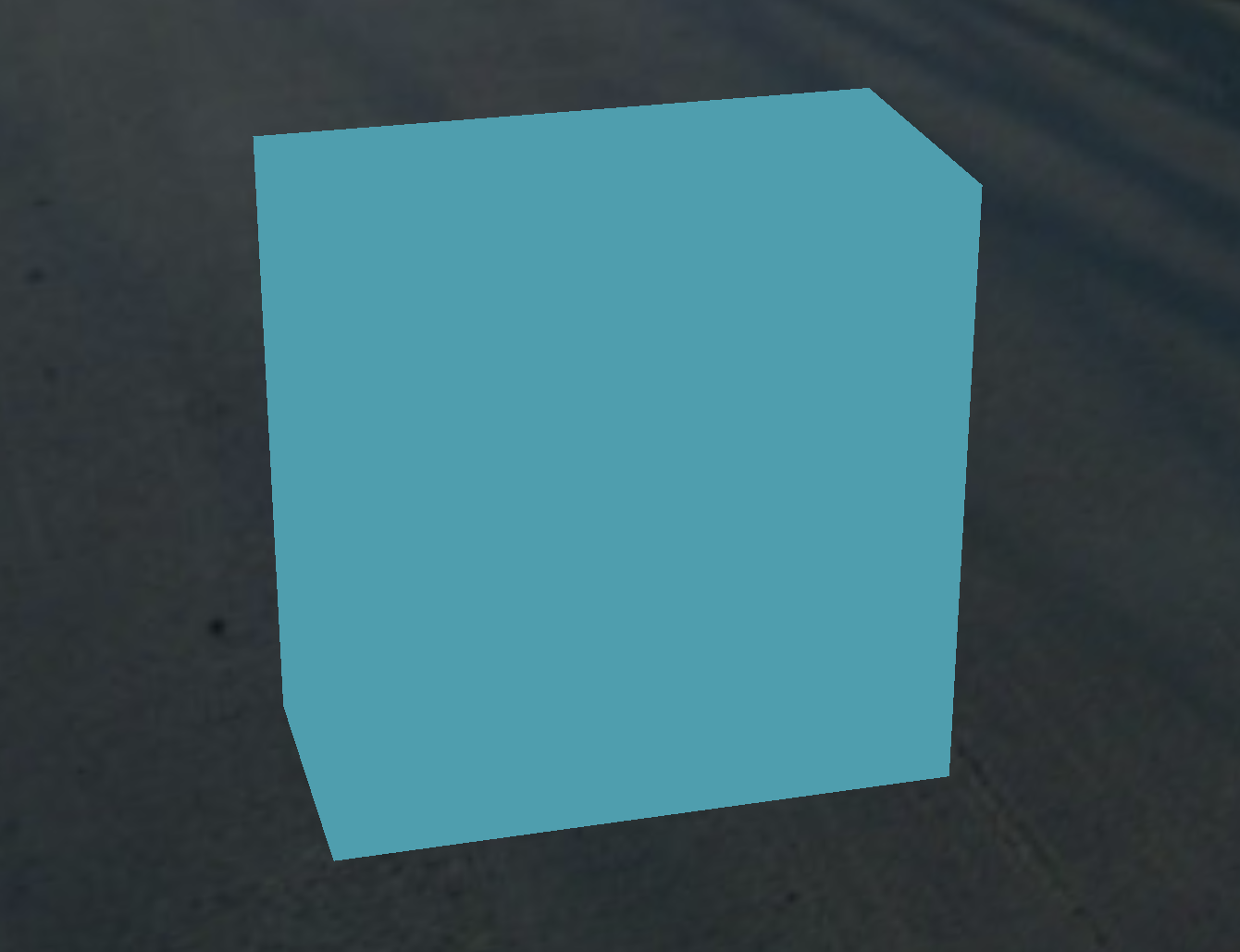}} \\
cylinder1   & Thick cylinder      & $D7 \times 15$          & 140 & --                              & \raisebox{-0.3\height}{\includegraphics[width=0.6cm]{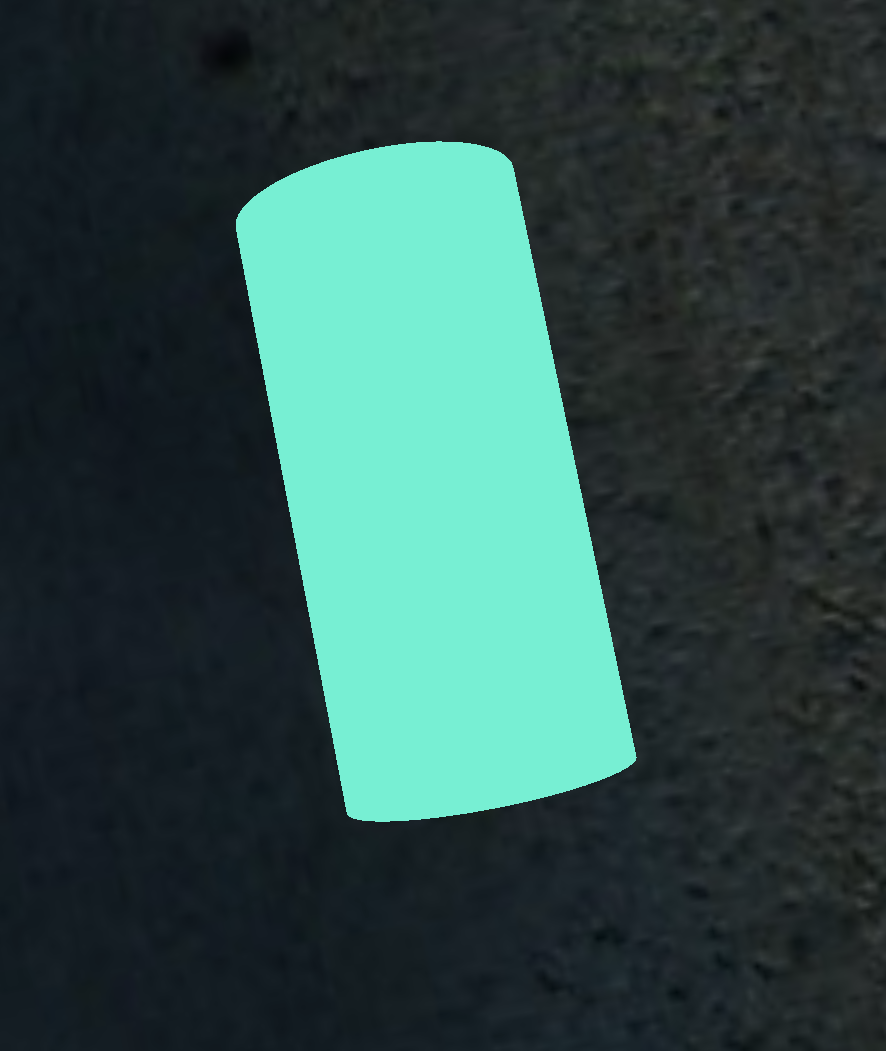}} \\
cylinder2   & Disk cylinder       & $D10 \times 2$          & 50  & --                              & \raisebox{-0.3\height}{\includegraphics[width=0.6cm]{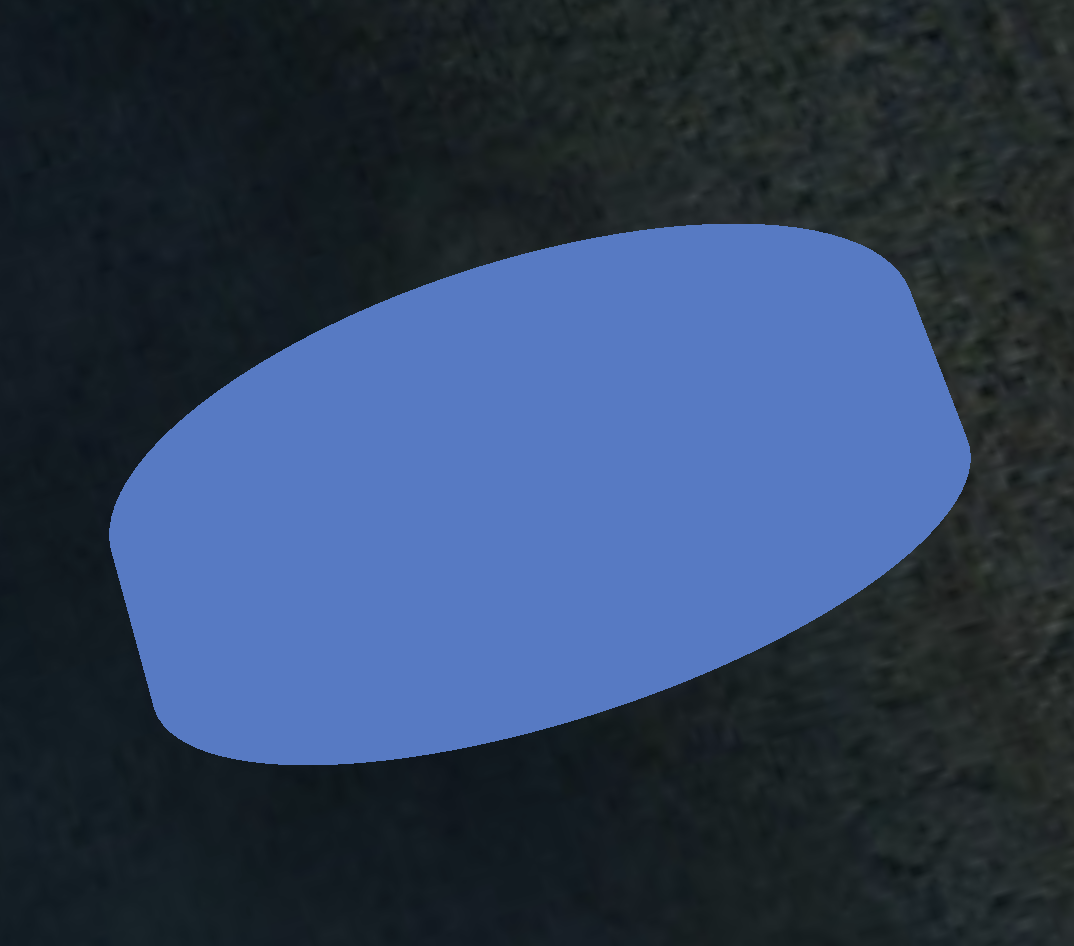}} \\
cylinder3   & Long cylinder       & $D3 \times 10$          & 25  & --                              & \raisebox{-0.3\height}{\includegraphics[width=0.6cm]{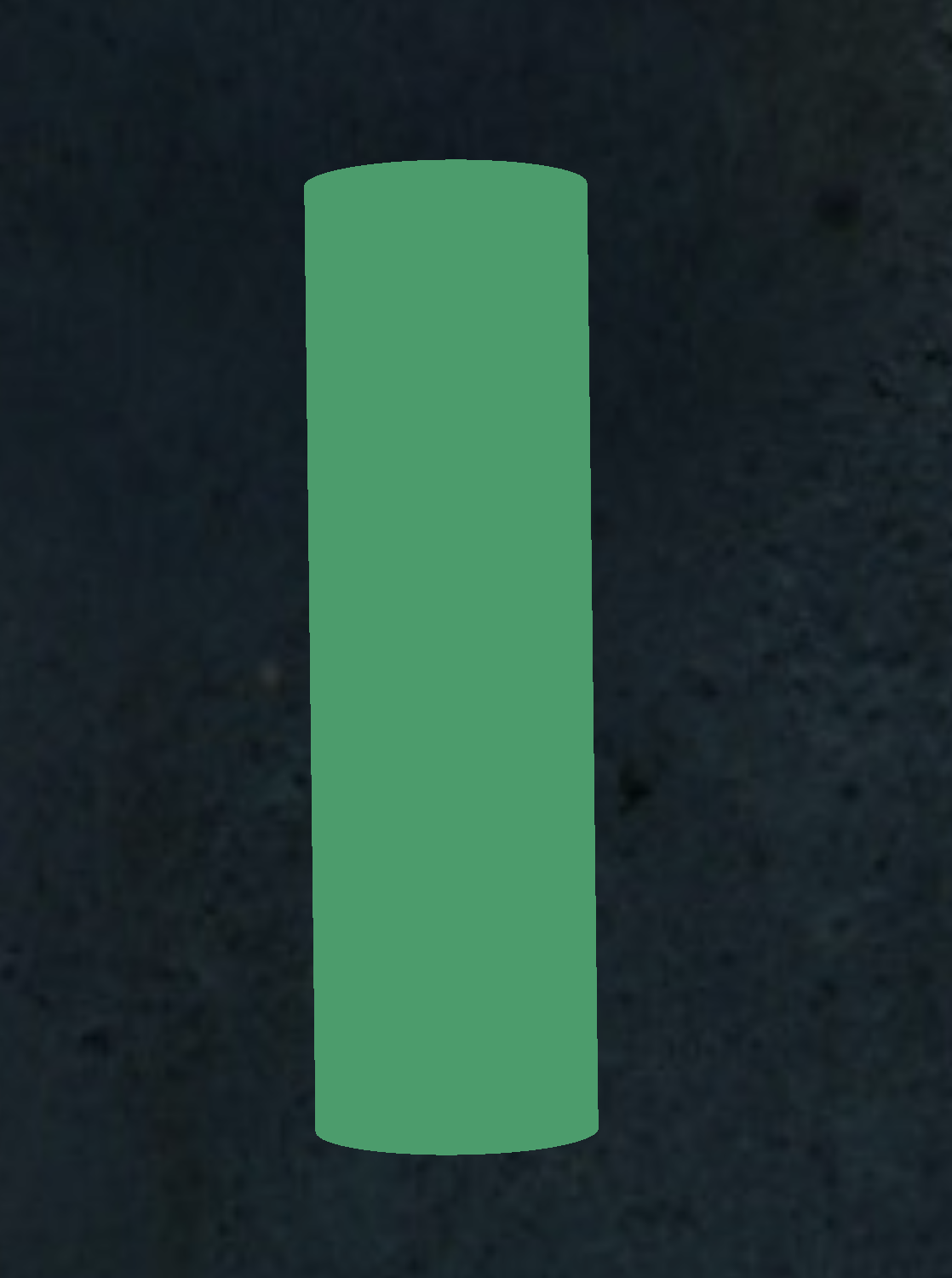}} \\
cylinder4   & Thin cylinder       & $D1 \times 15$          & 8   & --                              & \raisebox{-0.3\height}{\includegraphics[width=0.6cm]{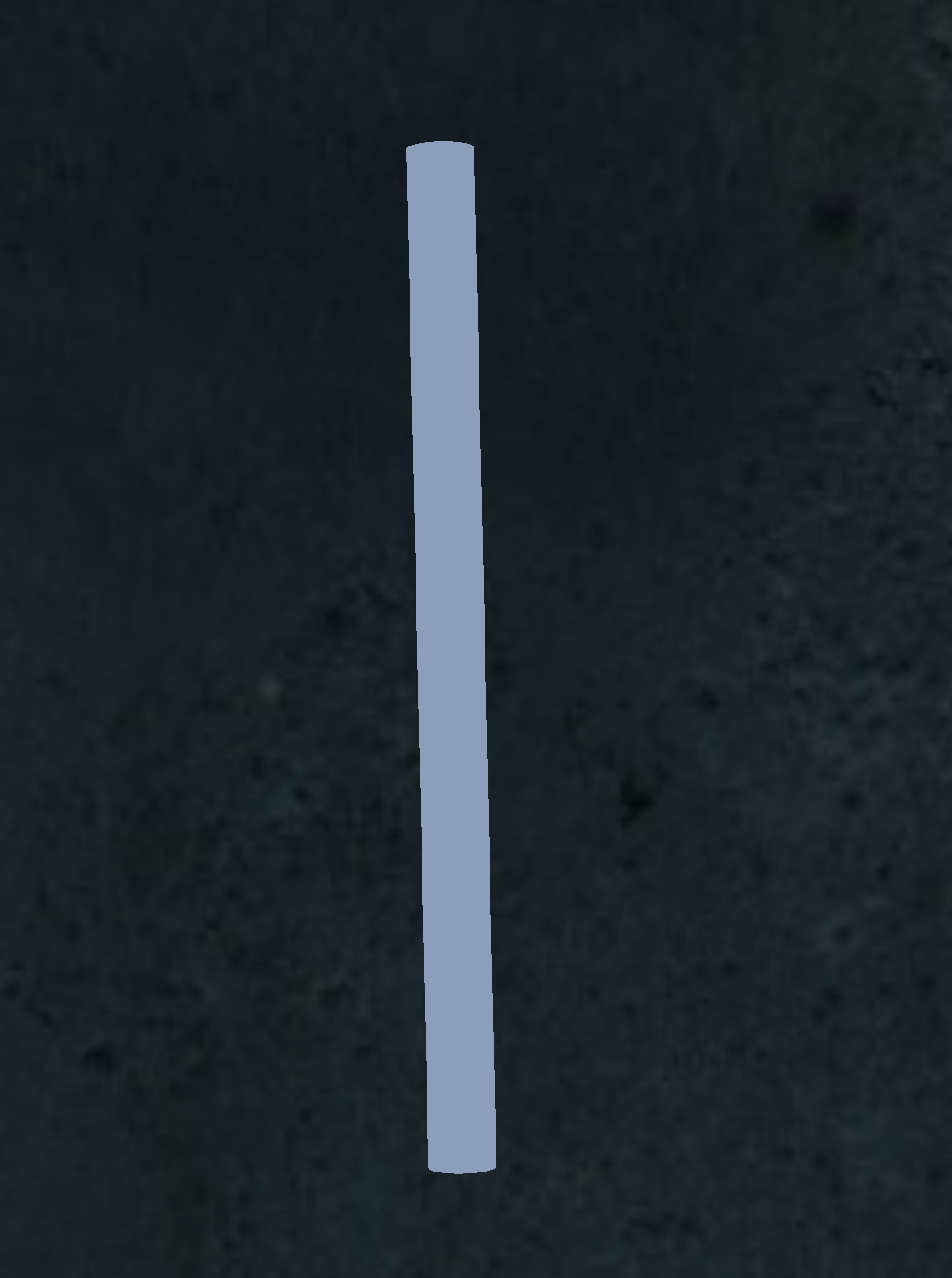}} \\
cylinder5   & Mini cylinder       & $D2 \times 2.7$         & 6   & --                              & \\
cylinder6   & Short thick cylinder& $D8 \times 10$          & 123 & --                              & \raisebox{-0.3\height}{\includegraphics[width=0.6cm]{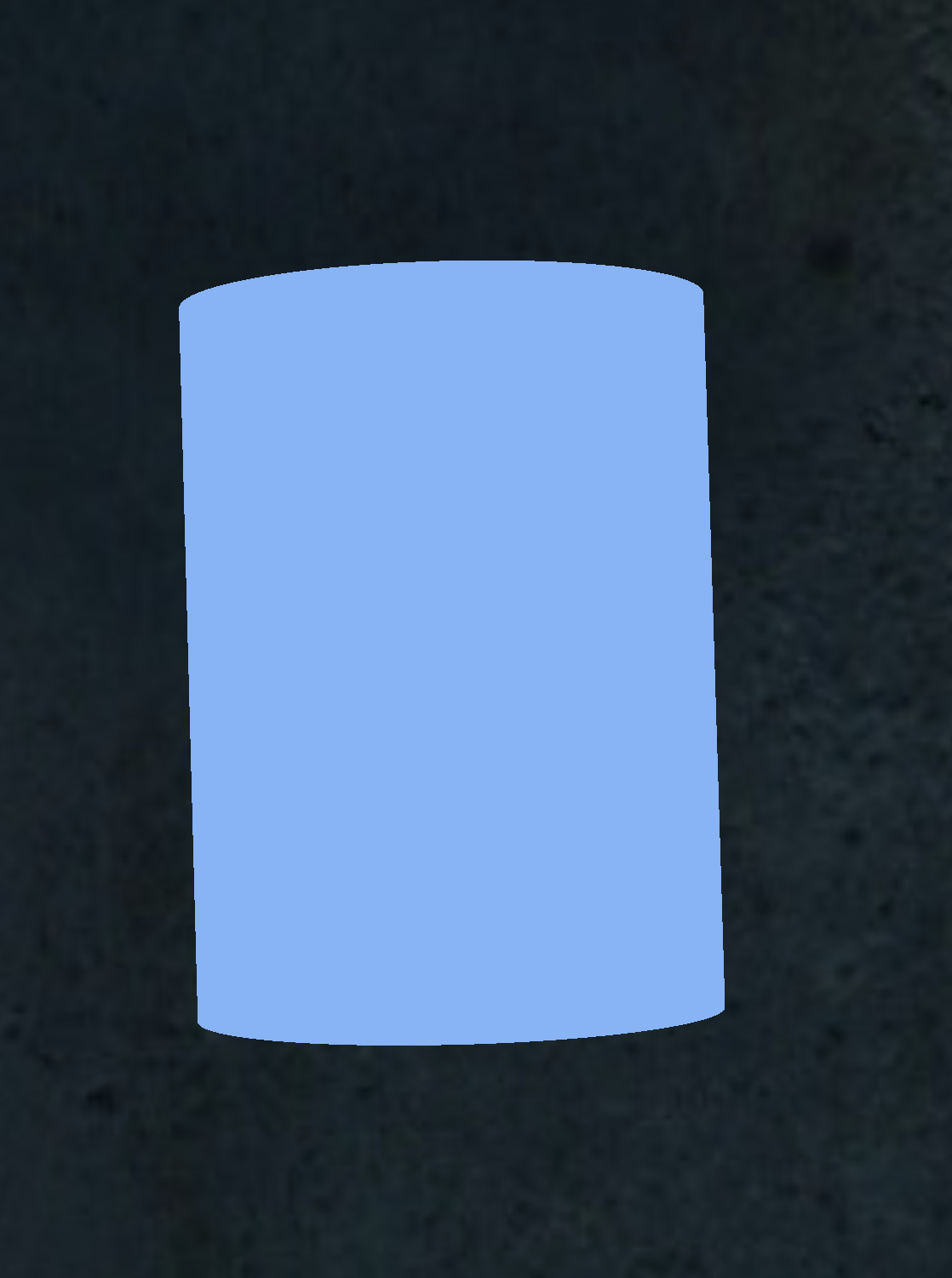}} \\
sphere1     & Large sphere        & $D4$                    & 13  & --                              & \raisebox{-0.3\height}{\includegraphics[width=0.6cm]{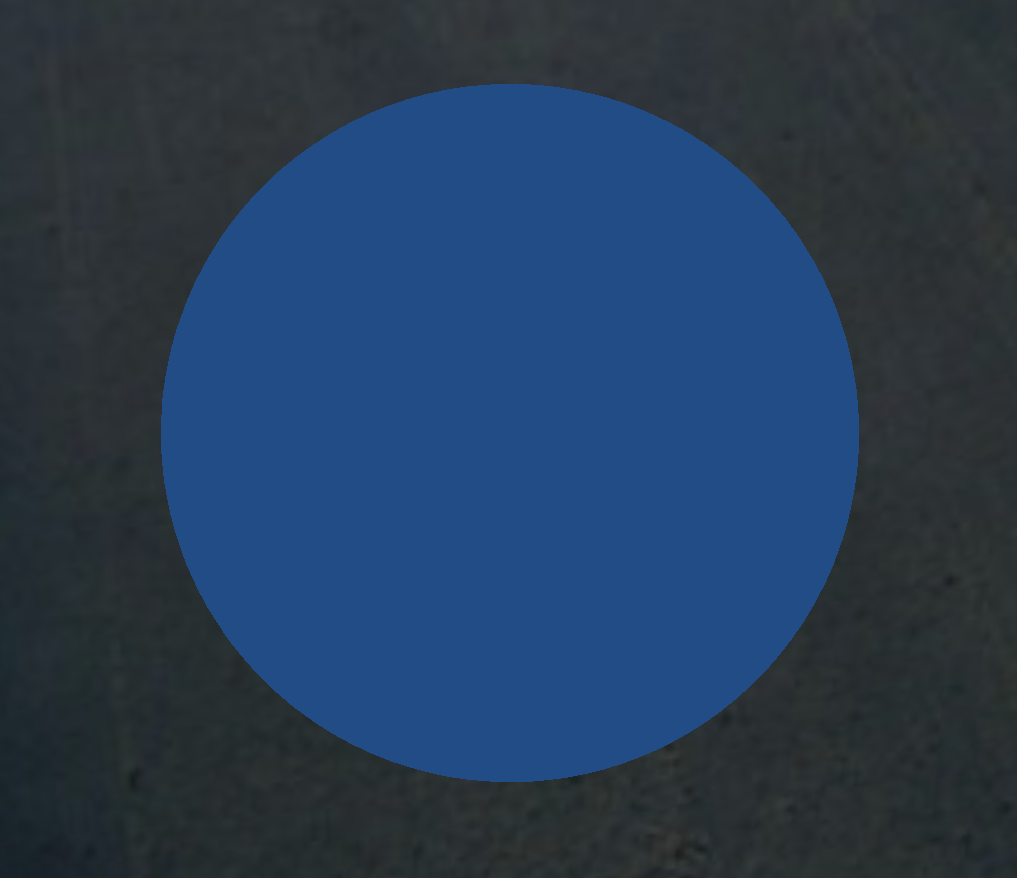}} \\
sphere2     & Medium sphere       & $D3$                    & 8   & --                              & \raisebox{-0.3\height}{\includegraphics[width=0.6cm]{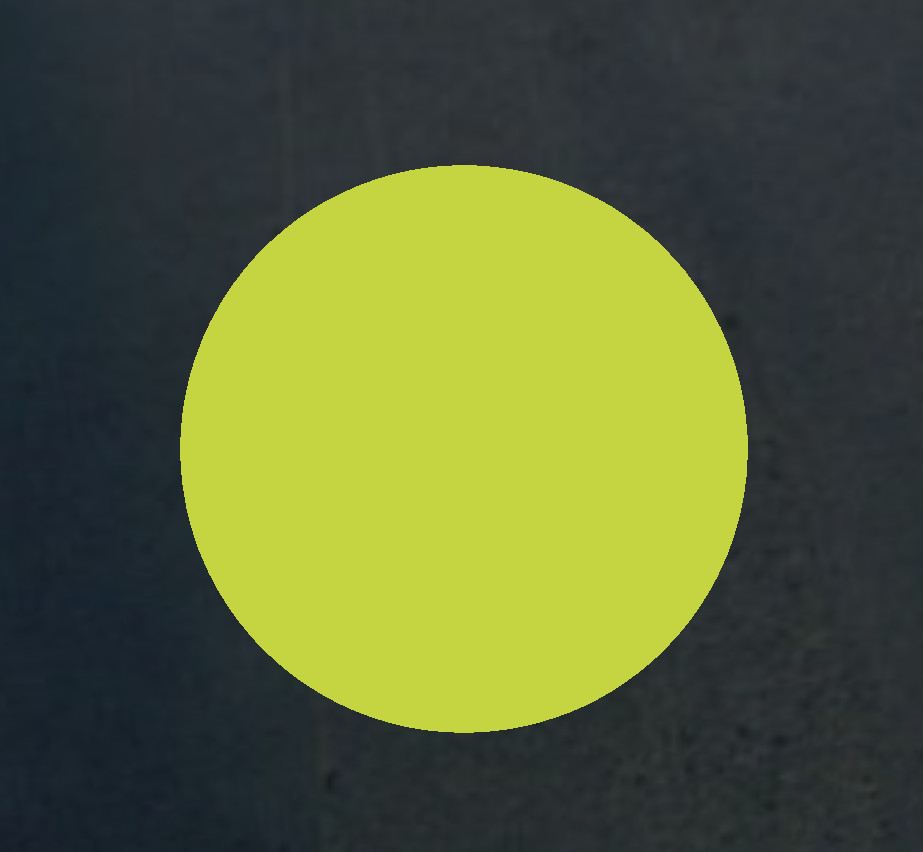}} \\
sphere3     & Small sphere        & $D2$                    & 4   & --                              & \raisebox{-0.3\height}{\includegraphics[width=0.6cm]{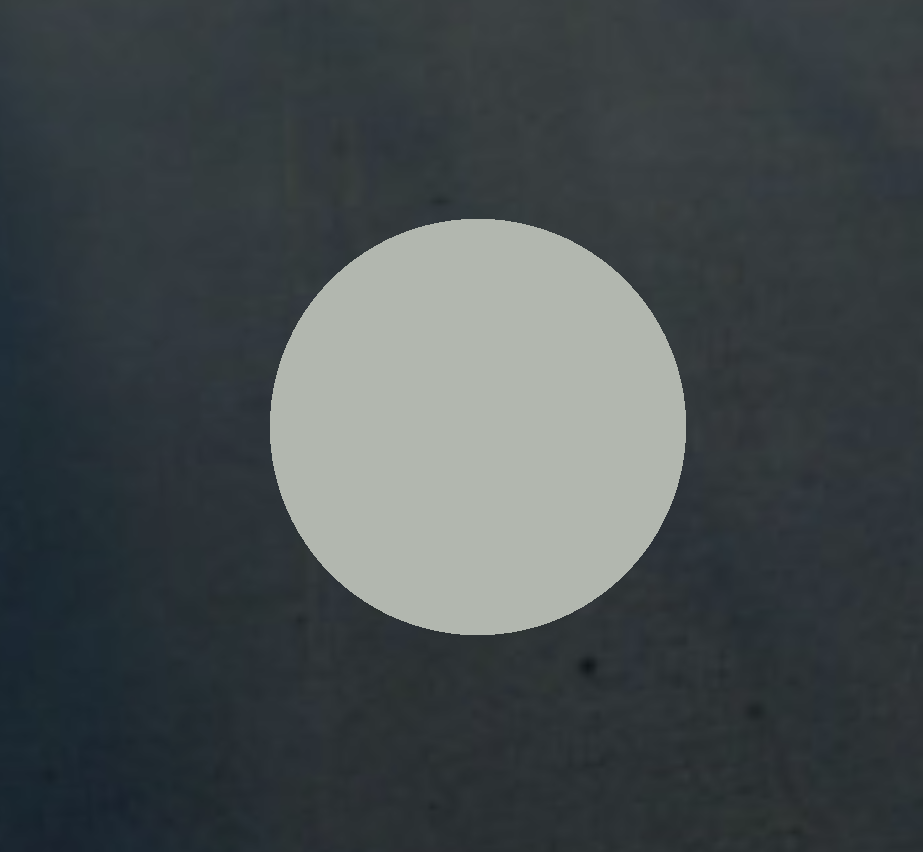}} \\
\midrule
cube2H      & Small cube (heavy)  & $5 \times 5 \times 5$   & 123 & --                              & \\
cuboid2H    & Long cuboid (heavy) & $15 \times 5 \times 2.5$& 203 & Long edge along $x$-axis        & \\
cuboid3H    & Small cuboid (heavy)& $5 \times 5 \times 2.5$ & 82  & --                              & \\
cylinder1H  & Thick cylinder (heavy) & $D7 \times 15$       & 554 & --                              & \\
cylinder4H  & Thin cylinder (heavy)  & $D1 \times 15$       & 14  & --                              & \\
cylinder5H  & Mini cylinder (heavy)  & $D2 \times 2.7$      & 10  & --                              & \\
cylinder6H  & Short thick cylinder (heavy) & $D8 \times 10$ & 472 & --                              & \\
sphere1H    & Large sphere (heavy) & $D4$                   & 35  & --                              & \\
\midrule
mayonnaisebottle & Mayonnaise bottle & -- & 224 & Base down              & \raisebox{-0.3\height}{\includegraphics[width=0.6cm]{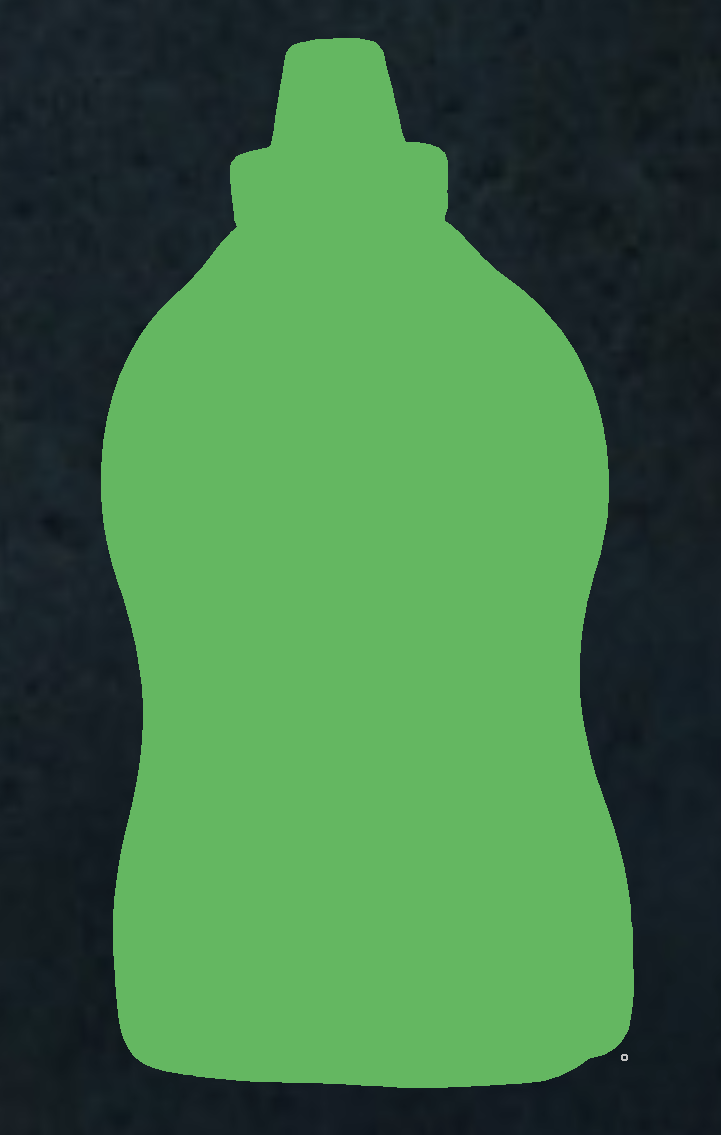}} \\
banana      & Banana              & -- & 75  & Long edge along $x$-axis & \raisebox{-0.3\height}{\includegraphics[width=0.6cm]{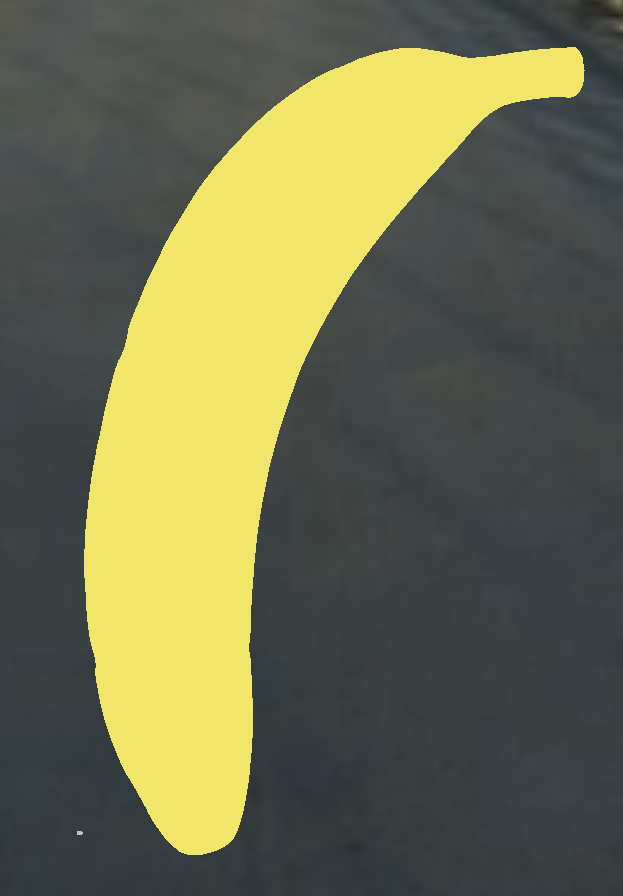}} \\
bowl        & Bowl                & -- & 42  & Base down              & \raisebox{-0.3\height}{\includegraphics[width=0.6cm]{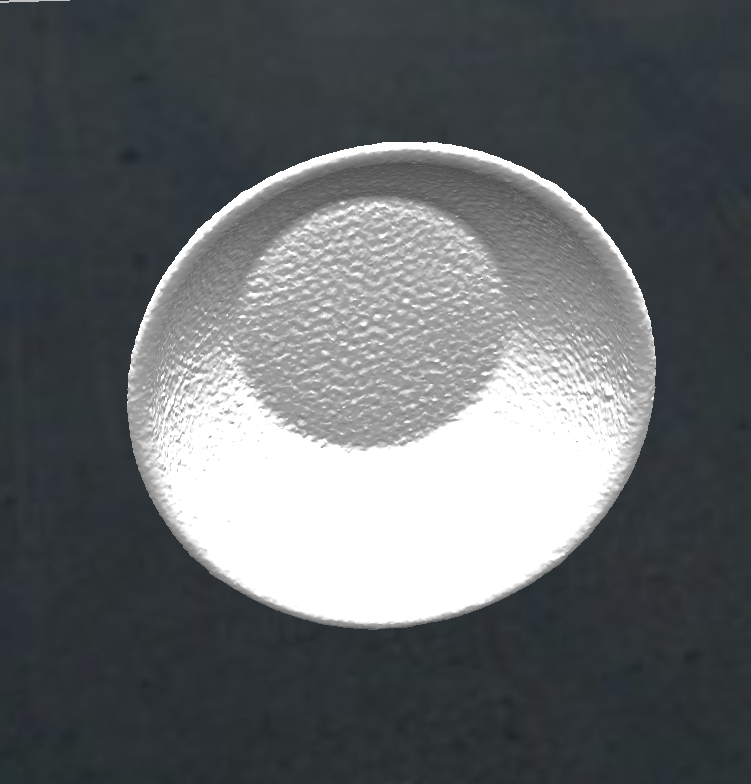}} \\
largeclamp  & Clamp               & -- & 59  & Base down              & \raisebox{-0.3\height}{\includegraphics[width=0.6cm]{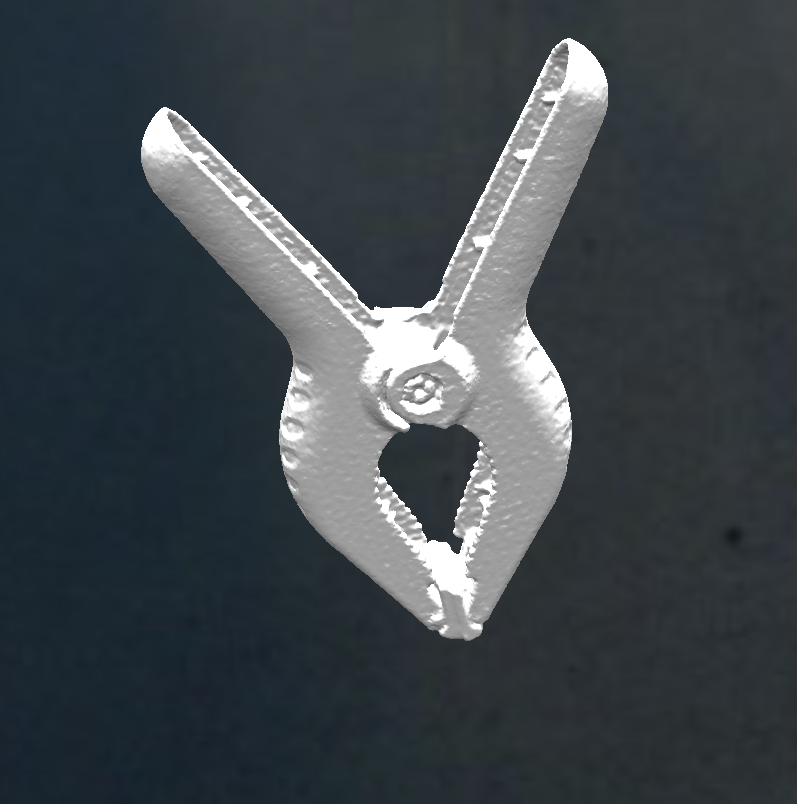}} \\
pitcherbase & Pitcher             & -- & 482 & Base down              & \raisebox{-0.3\height}{\includegraphics[width=0.6cm]{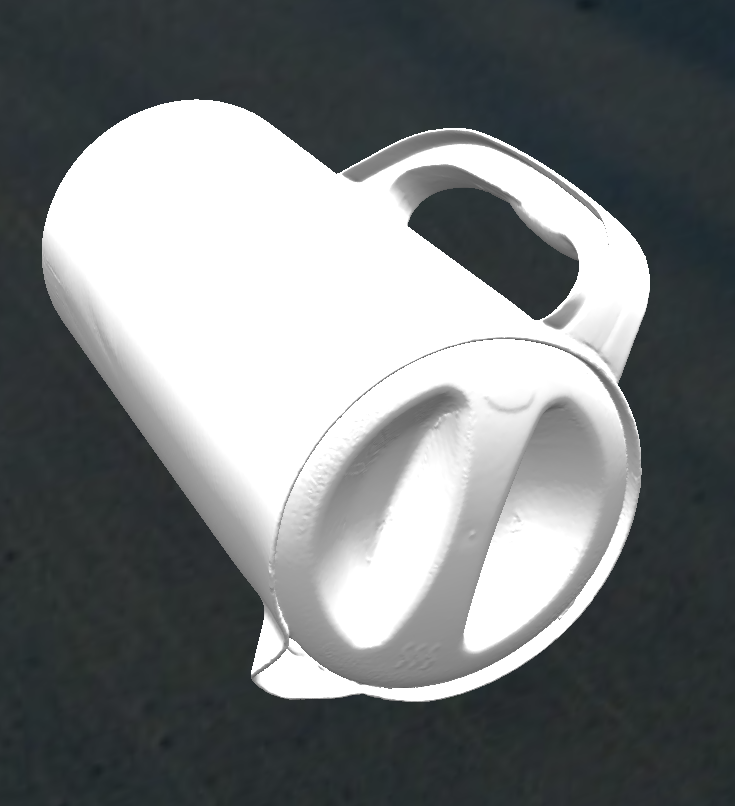}} \\
powerdrill  & Power drill         & -- & 228 & Handle along $z$-axis  & \raisebox{-0.3\height}{\includegraphics[width=0.6cm]{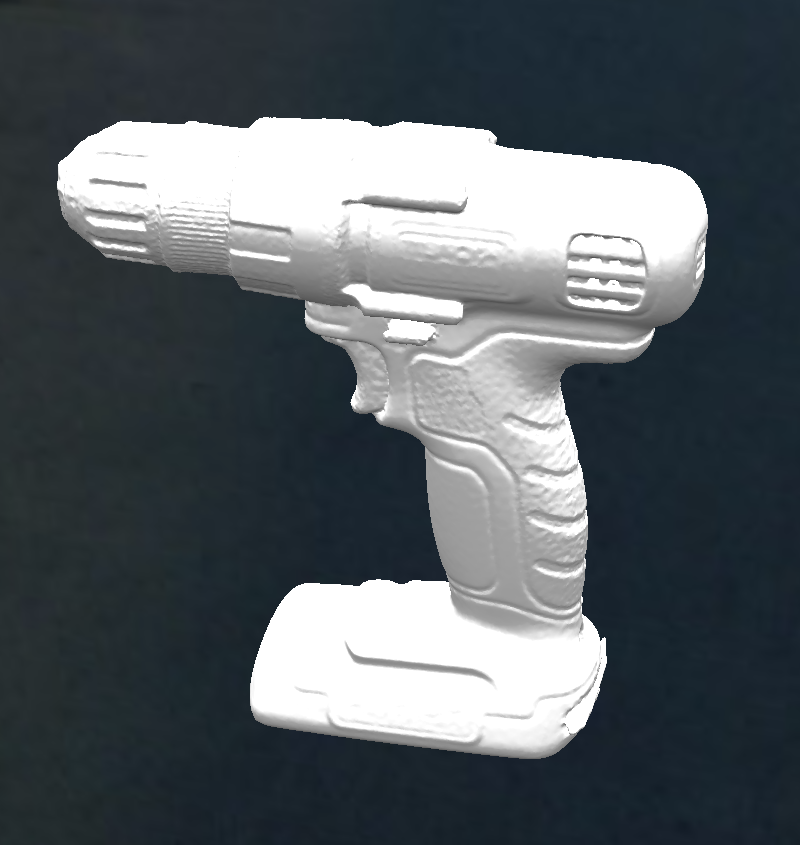}} \\
scissor     & Scissors            & -- & 32  & Base down              & \raisebox{-0.3\height}{\includegraphics[width=0.6cm]{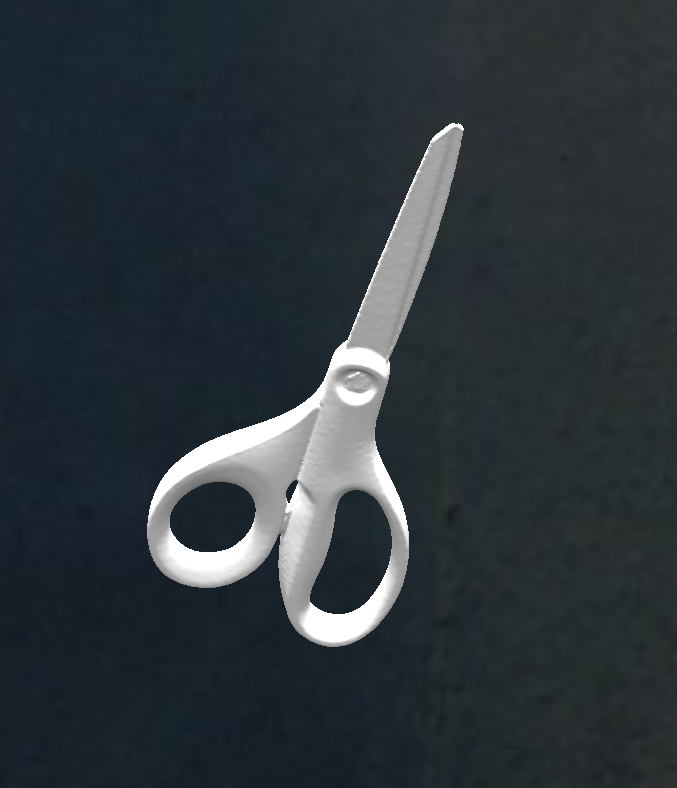}} \\
\bottomrule
\end{tabularx}
\end{table}

\section{Motion-Capture Processing}
\label{app:mano}

\textbf{Pre-processing.} We remove unusable segments, such as long occlusions or broken marker constellations, and tag anomalous trials. Each object's 6-DoF pose in the motion-capture world frame is recovered from its four markers by rigid-body registration. For the hand, subject-specific calibration fixes the marker-to-bone offsets, and the right-hand marker constellation is tracked to obtain the wrist pose and absolute joint rotations. A fixed transform $T_{MW}$ from the motion-capture world (W) to the MANO world (M) maps every pose:
\begin{equation}
T_H^{M}(t) = T_{MW}\,T_H^{W}(t), \qquad
T_O^{M}(t) = T_{MW}\,T_O^{W}(t), \qquad
R_j^{M}(t) = R_{MW}\,R_j^{W}(t),
\end{equation}
where $T_H^{W}(t)$ and $T_O^{W}(t)$ are the hand and object poses and $R_j^{W}(t)$ is the absolute rotation of joint $j$. All streams are time-aligned with the RGB-D cameras through recorded timestamps, and a short calibration gesture verifies synchronization so that each RGB-D frame corresponds to a well-defined hand--object configuration.

\textbf{MANO shape.} For each participant, we detect the right hand in the RGB frames of a calibration sequence, crop it, and run HaMeR~\citep{pavlakos2024reconstructing} to estimate MANO parameters. Averaging the shape estimates across crops with outlier rejection yields a stable shape vector $\hat\beta$ per subject. The shape is reused across sessions and in simulation.

\textbf{MANO fitting.} MANO maps the shape $\beta\in\mathbb{R}^{10}$, the pose $\theta$, and a global wrist transform to a differentiable hand mesh and joint locations via linear blend skinning. With $\hat\beta$ fixed and the marker-to-model correspondence calibrated, we solve Equation~\eqref{eq:mano_fit} for every frame, yielding world-frame wrist rotation and translation and local rotations for all MANO joints. Running the MANO forward model on $(\hat\beta,\theta_t,T_t)$ gives per-frame joints and mesh vertices, aligned to the motion-capture world frame and synchronized with the object pose and RGB-D streams.

\section{Reconstruction Policy Details}
\label{app:policy}

\textbf{MDP.} Each demonstration defines a Markov decision process $\mathcal{M}=(\mathcal{S},\mathcal{A},\mathcal{P},R,\gamma)$. The state $s_t$ contains hand and object kinematics together with the object and manipulation identifiers; the action $a_t\in\mathbb{R}^n$ specifies continuous residual increments; $\mathcal{P}$ is the simulated dynamics. The stochastic policy $a_t \sim \pi_\theta(a_t \mid s_t)$ maximizes the expected discounted return $J(\theta) = \mathbb{E}_{\pi_\theta}\big[\sum_{t=0}^{T}\gamma^{t} R(s_t,a_t)\big]$, with episodes ending on completion or failure within horizon $T$.

\textbf{Actions.} Table~\ref{tab:act} lists the action space. Equation~\eqref{eq:action_filter} turns raw actions into an exponentially weighted sum, $u_t = \sum_{i=1}^{t} \tau^{t-i} a_i$, so recent actions weigh most. For example, with $\tau = 0.9$, five identical wrist commands of 1\,cm along $x$ accumulate to $u_5 \approx 4.1$\,cm through a smooth ramp instead of discrete jumps, which keeps grasps from slipping under abrupt changes.

\begin{table}[H]
\centering
\caption{Action space. Actions are continuous and bounded.}
\label{tab:act}
\footnotesize
\begin{tabular}{@{}llcll@{}}
\toprule
\textbf{Group} & \textbf{Symbol} & \textbf{Dim.} & \textbf{Meaning} & \textbf{Bounds} \\
\midrule
Wrist translation & $a^{\text{lin}}_t$ & $3$ & Cartesian increments $(x,y,z)$ & $[-a_{\max},a_{\max}]$ m/step \\
Wrist rotation & $a^{\text{rot}}_t$ & $3$ & Roll--pitch--yaw increments & $[-r_{\max},r_{\max}]$ rad/step \\
Finger joints & $a^{\text{fin}}_t$ & $n_f$ & Per-joint increments (thumb and four fingers) & Per joint group \\
\midrule
Total & & $6+n_f$ & & \\
\bottomrule
\end{tabular}
\end{table}

\textbf{Observations.} Table~\ref{tab:obs} lists the observation. The raw object point cloud contributes 192 of its 470 dimensions; a PointNet encoder compresses it into a 64-dimensional feature, giving a 342-dimensional policy input.

\begin{table}[H]
\centering
\caption{Observation components of the reconstruction policy.}
\label{tab:obs}
\footnotesize
\begin{tabularx}{\linewidth}{@{}llc>{\raggedright\arraybackslash}X>{\raggedright\arraybackslash}X@{}}
\toprule
\textbf{Block} & \textbf{Symbol} & \textbf{Dim.} & \textbf{Contents} & \textbf{Notes} \\
\midrule
Wrist state & $x^w_t$ & $6$ & Wrist translation and rotation & Reference-frame hand state \\
Finger joints & $q_t$ & $20$ & Finger joint positions & Normalized by joint limits \\
Hand pose & $x^h_t$ & $7$ & Palm position and orientation & Position + quaternion \\
Fingertips & $x^{\text{tip}}_t$ & $15$ & Five fingertip positions & Object-relative \\
Object pose & $x^o_t$ & $7$ & Object position and orientation & Position + quaternion \\
Reference object pose & $\hat{x}^o_t$ & $7$ & Motion-capture object pose & Reference trajectory \\
Short-horizon target & $\hat{p}^o_{t+k}$ & $3$ & Future reference object position & Trajectory preview \\
Residual history & $(\Delta p_t,\Delta q_t)$ & $23$ & Accumulated position and joint residuals & Residual controller state \\
Object point cloud & $P_o$ & $64\times3$ & Object surface points & PointNet-encoded \\
Task identity & $c^{\text{act}}_t$ & $50$ & One-hot manipulation type & Manipulation intent \\
Object geometry & $g_o$ & $12$ & Compact shape descriptor & Type and size \\
Hand keypoints & $K^h_t$ & $16\times3$ & Hand keypoint positions & Object-relative \\
Contact forces & $f^c_t$ & $16$ & Contact force magnitudes & Tanh-normalized \\
Contact directions & $d^c_t$ & $16\times3$ & Contact force directions & Unit vectors; zero without contact \\
Expected contacts & $m^c_t$ & $16$ & Keypoints expected in contact & From the demonstration \\
\midrule
Total & $o_t$ & $470$ & \multicolumn{2}{l}{Raw observation before PointNet encoding} \\
\bottomrule
\end{tabularx}
\end{table}

\textbf{Rewards.} Table~\ref{tab:rewards} defines the reward components.

\begin{table}[H]
\centering
\small
\caption{Reward components for physically valid trajectory reconstruction.}
\label{tab:rewards}
\begin{tabularx}{\linewidth}{@{}llX@{}}
\toprule
\textbf{Component} & \textbf{Symbol} & \textbf{Definition} \\
\midrule
Position tracking & $r_{\text{pos}}$ & $\sum_{i\in\{x,y,z\}} \alpha_i \exp(-\lambda |p^i_{o,t}-\hat p^i_{o,t}|)\cdot g_t$ \\
Rotation tracking & $r_{\text{rot}}$ & $s_r \phi(\theta_t)+b_r$, with $\phi$ a three-segment piecewise function of the rotation error $\theta_t$ \\
Contact & $r_{\text{con}}$ & Coverage of the expected contacts within the contact-active window \\
Object stability & $r_{\text{stab}}$ & $r_{\max}^{\text{stab}}\exp[-(\|v_{o,t}\|/v_{\text{ref}})^2]$ after the contact window \\
Action penalty & $c_{\text{act}}$ & L1 penalty on the accumulated wrist and joint residuals $(\Delta p_t,\Delta q_t)$ \\
Survival & $r_{\text{surv}}$ & Small constant reward per valid step \\
Deviation penalty & $c_{\text{dev}}$ & Applied when the object deviates beyond the reset threshold \\
\midrule
Total & $r_t$ & $r_{\text{pos}}+r_{\text{rot}}+r_{\text{con}}+r_{\text{stab}}+r_{\text{surv}}-c_{\text{act}}-c_{\text{dev}}$ \\
\bottomrule
\end{tabularx}
\end{table}

\section{Diffusion Policy Configuration}
\label{app:diffusion}

\begin{table}[H]
\centering
\small
\caption{Diffusion Policy configuration for the experiments of Section~\ref{sec:dp_utility}.}
\label{tab:dp_config}
\begin{tabular}{@{}llr@{}}
\toprule
\textbf{Parameter} & \textbf{Symbol} & \textbf{Value} \\
\midrule
\multicolumn{3}{@{}l}{\textit{Problem setup}} \\
Observation dimension & $d_o$ & 508 \\
Action dimension & $d_a$ & 26 \\
Observation horizon & $T_o$ & 2 \\
Prediction horizon & $T_p$ & 16 \\
Executed action steps & $T_a$ & 8 \\
\midrule
\multicolumn{3}{@{}l}{\textit{Diffusion process}} \\
Training diffusion steps & $N_{\text{train}}$ & 100 \\
Inference denoising steps & $N_{\text{infer}}$ & 10 \\
\midrule
\multicolumn{3}{@{}l}{\textit{Network}} \\
U-Net channels & -- & $[128,256,512]$ \\
Diffusion-step embedding & $d_{\text{emb}}$ & 256 \\
Conv1D kernel size & $k$ & 5 \\
GroupNorm groups & $g$ & 8 \\
FiLM scale/bias conditioning & -- & yes \\
\midrule
\multicolumn{3}{@{}l}{\textit{Optimization}} \\
Batch size & $B$ & 256 \\
Learning rate & $\eta$ & $1\times10^{-4}$ \\
\bottomrule
\end{tabular}
\end{table}

\section{Retargeting Objective}
\label{app:retargeting}

Retargeting treats each frame as a numerical optimization over the robot's wrist pose and finger joint controls. Contact tokens extracted from the MANO hand--object interaction are mapped to the corresponding finger links of the target hand; each token defines a target contact point in the object frame. Forward kinematics gives the positions of the robot's contact candidates, and $E_{\text{target}}$ in Equation~\eqref{eq:retarget} is their distance to the target points. The remaining terms make the solution feasible and continuous: $E_{\text{pen}}$ penalizes hand--object penetration, $E_{\text{self}}$ self-collision, $E_{\text{joint}}$ joint-limit violation, and $E_{\text{temp}}$ changes between consecutive frames. Because the objective lives in the object frame and never compares joint angles across hands, the same formulation applies to any target hand whose links can be assigned to the demonstrated contacts.

\end{document}